\documentclass[pdflatex,sn-mathphys-num]{sn-jnl}

\usepackage{amsmath,amssymb,amsfonts}
\usepackage{algorithmic}
\usepackage{graphicx}
\usepackage{textcomp}
\usepackage{hyperref}
\usepackage{makecell}
\usepackage{caption}
\usepackage{float}
\usepackage{color}
\usepackage{subcaption}
\usepackage{multirow}
\usepackage{booktabs}
\usepackage{tikz}
\usetikzlibrary{3d,arrows.meta}

\begin{document}

\title[On the Separation of Human and AI-Generated Images in CLIP Embedding Space]{On the Separation of Human and AI-Generated Images in CLIP Embedding Space}
\title{On the Separation of Human and AI-Generated Images in CLIP Embedding Space}

\author[1]{\fnm{Andrea \sur{Asperti}}\email{andrea.asperti@unibo.it}}

\affil[1]{\orgdiv{Department of Informatics - Science and Engineering (DISI)}, \orgname{University of Bologna}, \orgaddress{\street{Via Mura Anteo Zamboni 7}, \city{Bologna}, \postcode{40126}, \country{Italy}}}

\abstract{
We identify a previously unreported phenomenon in CLIP representations:
human and AI-generated paintings spontaneously separate along the dominant
principal directions of their joint embedding distribution, without any
supervised objective designed to distinguish the two classes. Rather than
exploiting this phenomenon for detection, our objective is to interpret it:
we seek to identify the visual information underlying the separation and to
trace it back from the embedding space to the image domain.

We pursue this objective through a progressive investigation combining
interpretable image representations with gradient-based inversion, used
systematically as an experimental probe of the relationships identified in
feature space. Robustness experiments and increasingly expressive
statistical descriptors progressively rule out several intuitive
explanations based on global image properties and simple local statistics,
and point instead to distributed multiscale image structure. Multiscale
scattering provides the most informative interpretable representation
considered, but offers only a partial account of the phenomenon.

Direct inversion provides a complementary and striking observation:
substantial displacements along the dominant CLIP directions can be induced
by image perturbations that remain nearly imperceptible to human observers,
showing that the directions involved in the separation are highly sensitive
to image variations with very low perceptual salience for humans. Taken
together, these results reveal a significant difference between the visual
evidence reflected in CLIP representations and that readily accessible to
human perception, raising broader questions about the relationship between
artificial and human vision and, ultimately, between artificial and human
aesthetic judgment.}


\maketitle

\section{Introduction}
Several recent studies on AI-generated art \cite{artbench_dataset,Art_CLIPs_eyes} have shown that CLIP representations are highly effective at discriminating between human and AI-generated images, often outperforming non-expert human observers.
Remarkably, this phenomenon is common to different image generators, different collections of human paintings, and different versions of CLIP.

The most intriguing aspect of this phenomenon is that the separation does not need to be induced by supervised learning, but appears to emerge naturally from the geometry of the embedded data. More precisely, when the CLIP embeddings of a mixed collection of human and AI-generated paintings are projected onto a low-dimensional space through Principal Component Analysis (PCA), the two classes exhibit a substantial degree of separation along the dominant principal directions. 
This separation emerges without any optimization objective specifically designed for discrimination: the distinction between human and AI-generated paintings is already present in the empirical distribution of the embeddings.

\begin{figure}[ht]
    \centering
    \includegraphics[width=0.8\linewidth]{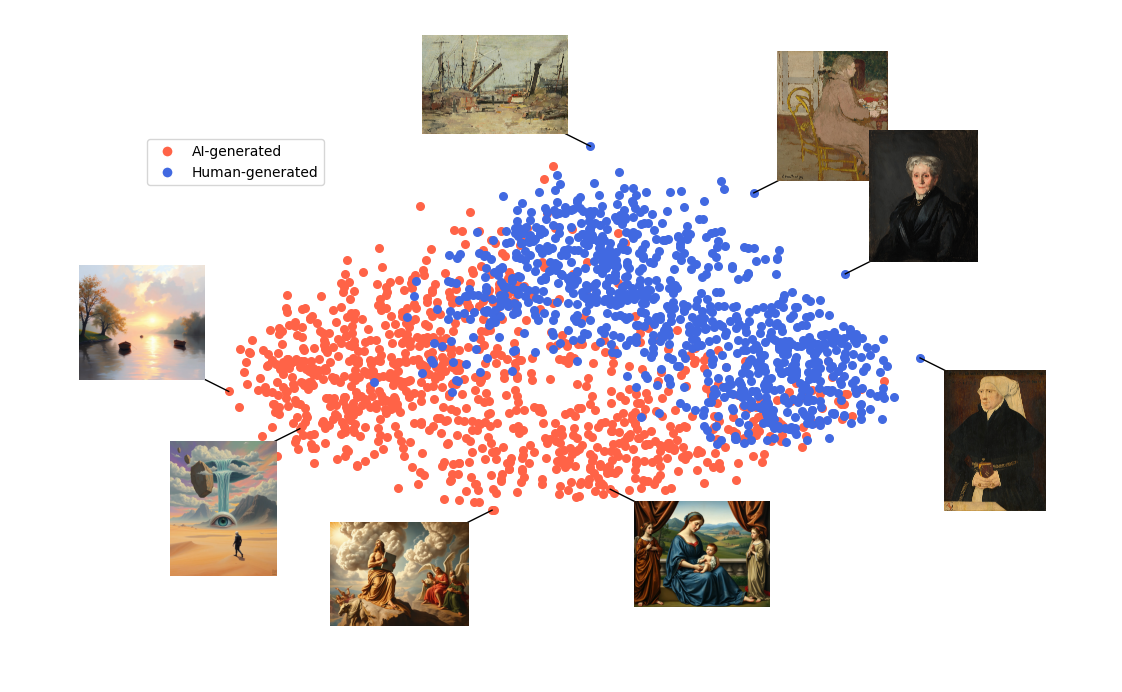}
    \caption{Two-dimensional PCA projection of CLIP embeddings for AI-generated paintings (red) and human paintings (blue).}
    \label{fig:PCA_projection}
\end{figure}

Figure~\ref{fig:PCA_projection} illustrates this phenomenon for images from the AI-Pastiche dataset \cite{AI-pastiche} (red points) and paintings from the National Gallery of Art, Washington \cite{nga_real_dataset} (blue points). The CLIP model employed throughout these experiments is \texttt{ViT-L/14@336px}.

The reasons underlying this phenomenon remain largely unexplained. Rather than developing a more accurate detector, our objective is to understand why this unsupervised separation emerges and to identify the visual information responsible for it. In particular, we seek to characterize

\begin{itemize}
\item interpretable image properties,
\item invariant statistical structures,
\item families of image transformations
\end{itemize}

that account for the dominant principal directions of the embedded dataset.

Our investigation follows a progressive strategy. We first establish that the observed separation is remarkably robust to a variety of transformations, including grayscale conversion, aggressive downsampling, cropping, and changes of image resolution. We then investigate increasingly expressive families of interpretable image representations. Elementary image statistics and local gradient descriptors explain only a limited fraction of the observed variability, allowing a number of intuitive hypotheses to be discarded. This progressive analysis ultimately leads to multiscale scattering representations, which provide the most informative statistical description of the phenomenon.

Empirically, the dominant principal directions are best explained by multiscale scattering features computed over image patches at different spatial resolutions, suggesting that the relevant information is encoded in distributed mesoscopic image statistics spanning multiple scales and locations rather than in isolated visual attributes.

Direct optimization experiments reveal an even more intriguing aspect of the phenomenon. By backpropagating through CLIP, it is possible to induce substantial displacements along the dominant principal directions through image perturbations that remain almost imperceptible to human observers (Section~\ref{sec:clip_inversion}). This observation provides an independent validation of the proposed statistical interpretation while revealing that CLIP relies on visual evidence largely inaccessible to human perception.

Taken together, these findings provide a partial explanation for the intrinsic geometry of CLIP embedding space. More broadly, they provide quantitative evidence of a systematic mismatch between artificial and human visual perception. Although both systems operate on the same paintings, they appear to rely on fundamentally different visual evidence. Our results suggest that CLIP exploits stable mesoscopic statistical regularities that remain largely inaccessible to visual inspection, whereas human observers predominantly rely on semantic, stylistic, and compositional properties. This discrepancy is particularly significant in the artistic domain, where visual perception constitutes the primary medium of evaluation. If humans and artificial vision systems rely on different visual evidence, the assumption that they share common aesthetic criteria becomes difficult to sustain. In this sense, the distinction between human and AI-generated paintings provides a unique experimental setting for investigating the differences between biological and artificial vision.

The main contributions of this work are the following:

\begin{itemize}

\item We identify and investigate a previously unexplained phenomenon: the spontaneous separation of human and AI-generated paintings along the dominant principal directions of CLIP embeddings.

\item We develop a progressive analytical framework based on increasingly expressive families of interpretable image representations, systematically testing and ruling out several plausible explanations for the observed separation.

\item We show that multiscale scattering statistics capture a significant, but still limited, fraction of the observed embedding geometry. This provides a partial explanation of the phenomenon and points to the relevance of distributed mesoscopic image structures, while leaving a substantial part of the separation unexplained.

\item Through gradient-based inversion, we show that substantial displacements along the dominant embedding directions can be induced by image perturbations that remain almost imperceptible to human observers. This provides direct evidence of a mismatch between the visual information exploited by CLIP and that accessible to human perception, with potential implications for computational aesthetics and the understanding of artificial vision.

\end{itemize}

The remainder of the paper follows the progressive investigative strategy outlined above. Section~\ref{sec:background} reviews the relevant background on CLIP representations, AI-generated image detection, interpretable image descriptors, and representation inversion. Section~\ref{sec:experimental_setting} introduces the experimental setting, including the datasets, CLIP models, and the inversion framework employed throughout the study. Section~\ref{sec:methodology} presents the general methodology. The core of the paper is organized as a progressive search for an explanation of the observed separation. 
In Section~\ref{sec:robustness}, the robustness of the
separation to a wide range of image transformations is investigated.
Sections~\ref{sec:global_statistics}-\ref{sec:scattering} progressively investigate increasingly expressive image representations, from elementary statistics and local gradient descriptors to multiscale scattering features.
%Section~\ref{sec:global_statistics} studies whether elementary image statistics account for the dominant principal directions and shows that they provide only limited explanatory power. Section~\ref{sec:HOG} extends the analysis to local gradient descriptors through HOG features, demonstrating that they capture part of the phenomenon but remain insufficient. Section~\ref{sec:scattering} introduces multiscale scattering representations, which provide the strongest statistical explanation of the embedding geometry. 
Section~\ref{sec:inversion} validates the results of the previous Sections through gradient-based inversion experiments, revealing image perturbations that substantially alter the CLIP representation while remaining nearly imperceptible to human observers. Finally, Section~\ref{sec:discussion} discusses the implications of these findings for the understanding of artificial vision and computational aesthetics, while Section~\ref{sec:conclusion} concludes the paper and outlines directions for future research.

\section{Background and Related Work}
\label{sec:background}
The present work combines ideas originating from several research areas, including interpretable image representations, representation inversion, and CLIP-based image analysis. This section briefly reviews the concepts most directly related to our approach.

\subsection{Understanding CLIP Representations}

Contrastive Language–Image Pretraining (CLIP) \cite{CLIP} has emerged as one of the most influential vision-language models, providing a shared embedding space in which images and text are represented through contrastive learning. Owing to its remarkable transferability, CLIP has become a standard visual representation for a wide range of downstream tasks, including image retrieval, recognition, segmentation, and multimodal reasoning. Beyond its practical success, its embedding space has attracted increasing attention as an object of study in its own right.

Several studies have shown that CLIP embeddings exhibit a rich semantic organization, in which high-level visual concepts and semantic attributes are encoded in structured regions of the embedding space and can be effectively aligned with natural language descriptions \cite{CLIP, splice24, goh2021multimodal}. Other works have investigated the internal organization of CLIP by analyzing the contribution of individual transformer layers, attention heads, and image patches to the final representation, revealing that semantic localization and concept-specific processing emerge naturally within the model \cite{CLIP_via_text24}. More recently, representation decomposition techniques have been proposed to express CLIP representations in terms of human-interpretable concepts, providing insight into the semantic structure of individual neurons and embedding components \cite{CLIP-dissect, splice24}.

While these studies have substantially improved our understanding of the semantic organization of CLIP representations, they provide limited insight into why particular image collections naturally separate within the embedding space.

\subsection{AI-Generated Image Detection}

The rapid diffusion of text-to-image models has stimulated extensive research on distinguishing AI-generated images from human-created ones. Early approaches relied on handcrafted forensic features or generator-specific artifacts \cite{easy_to_spot20,GragnanielloCMP21}, whereas more recent methods increasingly exploit large pretrained vision models, particularly CLIP, as generic feature extractors \cite{Raising_the_bar24, DetectingAI}. CLIP-based detectors have demonstrated remarkable robustness and generalization across generators, often requiring only lightweight classifiers trained on frozen embeddings \cite{Raising_the_bar24}.

Nevertheless, to the best of our knowledge, existing approaches formulate the problem exclusively as supervised classification.
The objective is to optimize a decision boundary that discriminates between real and synthetic images while maximizing classification accuracy and generalization to unseen generators. Recent work has also investigated robustness, cross-generator generalization, and interpretation of frozen CLIP detectors, but still within the supervised detection paradigm \cite{Improving_interpretability24,Raising_the_bar24}.

Rather than training a detector, we investigate the geometry of the embedded dataset itself, seeking to explain why the observed separation emerges before any supervised learning is performed.

\subsection{Interpretable Image Representations}

Long before the advent of deep learning, computer vision relied extensively on handcrafted image representations designed to capture well-defined geometric and statistical properties of natural images. These descriptors have been successfully employed in object recognition, texture analysis, image retrieval, and scene understanding, and remain valuable because of their mathematical interpretability and their ability to characterize specific visual phenomena. A comprehensive review of local image descriptors is provided by \cite{tuytelaars08local}, while \cite{statistics_for_images} provides an interesting survey of statistical approaches to image representation and analysis.

Among the most influential local descriptors are the Scale-Invariant Feature Transform (SIFT) \cite{lowe2004distinctive} and Histograms of Oriented Gradients (HOG) \cite{dalal2005histograms}, both of which characterize local image structure through gradient-based statistics while exhibiting robustness to moderate geometric and photometric variations. These representations have become standard tools for describing shape, edges, and local texture, providing features that are both computationally efficient and readily interpretable.

A complementary line of research models images through multiscale statistical representations derived from wavelet decompositions and natural image statistics. The seminal work of Portilla and Simoncelli \cite{portilla2000parametric} demonstrated that a large class of natural textures can be characterized through joint statistics of multiscale wavelet coefficients, while subsequent studies emphasized the broader role of natural image statistics in visual representation and perception \cite{simoncelli2001natural}. 
Building upon these ideas, the Scattering Transform \cite{mallat2011group, bruna2013invariant,sifre2013rotation} provides a mathematically grounded representation obtained by cascading wavelet convolutions, modulus nonlinearities, and local averaging operations. Scattering coefficients are stable to small deformations while preserving rich multiscale information about image structure, and have demonstrated excellent performance in texture discrimination and image classification. 

In the present work, HOG and Scattering are therefore employed not as alternatives to deep representations, but as analytical tools for interpreting the geometry of CLIP embeddings.

\subsection{Representation Inversion and Feature Visualization}

Understanding the visual information encoded by deep neural networks has motivated a broad family of visualization, attribution, and inversion techniques. Originally introduced to interpret learned representations and understand the hierarchical organization of deep features, inversion methods have recently experienced renewed interest owing to their central role in generative models, latent-space editing, and representation analysis.

Activation maximization methods synthesize inputs that strongly excite individual neurons or output classes, providing insight into the semantic concepts learned by deep representations \cite{erhan2009visualizing,yosinski2015understanding}. Complementary feature inversion techniques reconstruct images from intermediate representations, enabling the analysis of the information preserved at different stages of a neural network \cite{Vevaldi15,inverting_CNN}. More generally, optimization-based inversion has become an important tool for probing latent representations, investigating model behavior, and understanding the geometry of learned feature spaces.

Closely related are adversarial optimization methods, which compute small image perturbations capable of producing large changes in a network's internal representation or output \cite{Intriguing2013, goodfellow2014explaining}. While these techniques are typically employed to study the robustness and vulnerabilities of deep models, they also provide a powerful means of exploring the geometry of learned representations.

In the present work, inversion is employed as an analytical probe of the empirical geometry of the CLIP embedding space, rather than as a visualization or adversarial technique.

\subsection{Position of This Work}

Existing studies have investigated the semantic organization of CLIP representations, developed increasingly robust methods for AI-generated image detection, proposed interpretable handcrafted image representations, and employed inversion techniques to analyze learned models. However, these lines of research have largely evolved independently. To the best of our knowledge, no previous work has investigated why human and AI-generated paintings naturally separate in the empirical distribution of CLIP embeddings without supervised learning, nor related this phenomenon to interpretable image statistics through a combination of robustness analysis, progressively more expressive image descriptors, regression, and inversion. The present work addresses this gap.

\section{Experimental Setting} % Datasets, PCA, Inversion
\begin{figure}[ht]
    \centering
    \includegraphics[width=0.8\linewidth]{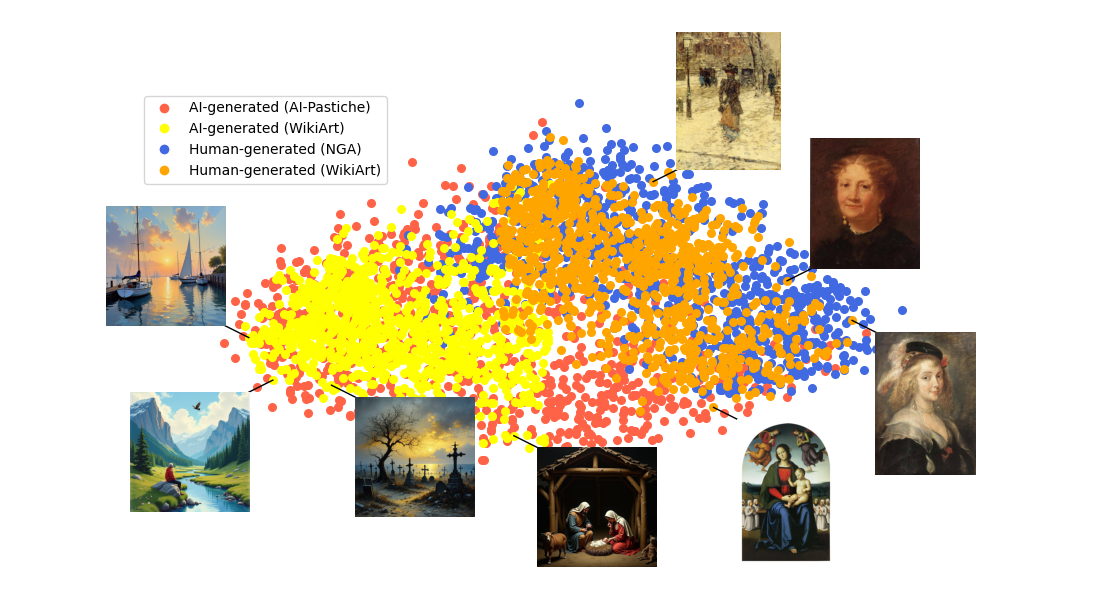}
    \caption{Two-dimensional PCA projection of CLIP embeddings for a random subset of AI-generated paintings (yellow) and human paintings (orange) borrowed from the AI-WikiArt dataset.
    }
    \label{fig:pca_wikiart}
\end{figure}

\label{sec:experimental_setting}
The phenomenon illustrated in Figure~\ref{fig:PCA_projection} is consistently observed across different datasets, image generators, and CLIP variants.
 To make an alternative example, in Figure~\ref{fig:pca_wikiart} we show the 2D PCA projection of one thousand human and ai generated images borrowed from the AI-WikiArt \cite{WikiArt25} dataset.

For the purposes of this article, we mostly worked with AI-WikiArt \cite{WikiArt25}, AIPastiche \cite{AI-pastiche}, and paintings from the National Gallery of Art Dataset (NGAD) \cite{nga_real_dataset}.

The AI-WikiArt dataset is a large-scale collection developed to evaluate vision-language models on tasks such as authorship attribution and AI-generated art detection \cite{WikiArt25}. It contains both authentic and synthetic artworks.
The real subset includes 39,530 WikiArt paintings from 128 identified artists, covering 10 genres and 27 artistic styles, with metadata for each image. The synthetic subset was created with three text-to-image models: Stable Diffusion, Flux, and F-Lite.

The AI-Pastiche dataset is a smaller dataset of generated paintings; however, it covers a much larger variety of generators, including very recent ones 
like Imagen4, FireflyImage4, Playgroundv2.5, Midjourney 7, Flux.2 and gpt-image-1.5. In its current version, AI-pastiche contains 1474 AI-generated paintings inspired by well-known artistic styles of the Western tradition, from the Renaissance onward. The dataset was constructed using 73 manually designed textual prompts, combined with 19 different image generators spanning multiple generations of diffusion-based and generative models. The dataset also includes extensive metadata documenting the generation process, including prompts, models, and generation settings. Additionally, it provides human based annotations, relative to perceived authenticity, prompt adherence, and defects. 

As an additional reference set of human artworks, we used the National Gallery of Art Dataset (NGA) \cite{nga_real_dataset}, obtained from the publicly available collection of the National Gallery of Art, Washington. The dataset contains European paintings together with associated metadata, including title, attribution, artistic period, and links to high-resolution images distributed through the IIIF protocol.

For our experiments, we mostly used AI-WikiArt for training, reserving AIPastiche and NGA for testing.

As for CLIP, the results presented in this article have mostly been 
obtained using the CLIP model \texttt{ViT-L/14@336px}, which rescales input images to a standard resolution of $336 \times 336$ pixels and produces embedding vectors of dimension $768$. Additional experiments with different 
CLIP versions didn't show any significant differences.

\subsection{2D PCA Projection}
As already observed in the introduction, the striking
phenomenon is not the possibility to train a supervised
classifier to discriminate between Human and AI-generated artworks, but the fact that this distinction manifests itself
in a completely natural way when considering the principal
components of the combined dataset, as shown in 
Figure~\ref{fig:PCA_projection}. In fact, more than explaining the separability between the two classes, we
are even more interested in elucidating the variation factors 
that govern the movement along the two main components 
PC1 and PC2 of the 2D projection.

In explicit terms, the goal of our research can be further refined to:
\begin{itemize}
    \item explain what PC1 and PC2 correspond to in image space; 
    \item identify image modifications capable of reducing the observed separation.
\end{itemize}

In the rest of the article, we shall frequently project images in the 2D space defined by the original PC1 and PC2,
to understand how different image transformations are reflected along these projections. An example will be
given in the next section, where we invert PC1 through 
CLIP in real space and reproject the transformed image
back into the 2D space, to observe its displacement.

Finally, it is worth noting that, before PCA projections, the two sets of pictures (AI and Human) are relatively close in CLIP's embedding space, as conceptually exemplified in Figure \ref{fig:example}

\begin{figure}[ht]
    \centering
    %\documentclass[tikz,border=5pt]{standalone}
%\usepackage{tikz}
%\usetikzlibrary{3d,arrows.meta}

%\begin{document}

\begin{tikzpicture}[scale=3, >=Latex]

% sphere outline
\draw[gray!60, thick] (0,0) circle (1);

% latitude / longitude hints
\draw[gray!30] (-1,0) arc (180:360:1 and 0.28);
\draw[gray!30,dashed] (1,0) arc (0:180:1 and 0.28);

\draw[gray!30] (0,-1) arc (-90:90:0.28 and 1);
\draw[gray!30,dashed] (0,1) arc (90:270:0.28 and 1);

% red cluster
\foreach \p in {
  (0.45,0.58),(0.50,0.52),(0.40,0.65),(0.57,0.60),
  (0.48,0.70),(0.35,0.75),(0.52,0.64),(0.42,0.49),
  (0.34,0.56),(0.30,0.62)
}{
  \fill[red] \p circle (0.018);
}

% blue cluster
\foreach \p in {
  (0.55,0.45),(0.62,0.38),(0.48,0.40),(0.58,0.53),
  (0.68,0.47),(0.52,0.33),(0.65,0.29),(0.55,0.40),
  (0.59,0.27),(0.69,0.40)
}{
  \fill[blue] \p circle (0.018);
}

% centroid vectors
\draw[red, very thick, ->] (0,0) -- (0.45,0.65);
\draw[blue, very thick, ->] (0,0) -- (0.63,0.45);

% labels
\node[red] at (0.50,0.82) {AI};
\node[blue] at (0.70,0.15) {Human};

\node[gray!70] at (0,-1.15) {normalized embedding sphere};

\end{tikzpicture}

%\end{document}
    \caption{Conceptualization of the relative position of the manifolds of human and AI-generated paintings in CLIP's embedding space.}
    \label{fig:example}
\end{figure}

The previous observation is important for understanding that the components we are interested in are not general components of CLIP's embedding space, but specific components of our manifold of data.

\subsection{Direct CLIP Inversion and Perceptual Dissociation}
\label{sec:clip_inversion}
To better understand the geometric structure underlying the observed separation, we performed direct optimization experiments inside CLIP space. Starting from an AI-generated image $x_0$, we searched for a perturbation capable of moving the image along selected principal directions of the CLIP embedding space while preserving visual similarity from the point of view of a human observer.

More precisely, let
\[
f(x)\in\mathbb{R}^{768}
\]
denote the normalized CLIP embedding of an image $x$, and let
\[
v_1,v_2
\]
be the first principal directions obtained from PCA over the full embedding dataset. The corresponding PCA coordinates are computed as
\[
p_i(x)=\langle f(x)-\mu,v_i\rangle,
\]
where $\mu$ denotes the empirical mean embedding.

The optimization problem consists in finding a perturbed image
\[
x=x_0+\delta
\]
that maximizes or minimizes one of the principal coordinates while remaining visually close to the original image.

The perturbation is parameterized as
\[
\delta=\varepsilon\tanh(u),
\]
where $u$ is the optimization variable and $\varepsilon$ controls the maximum perturbation amplitude. This parameterization ensures bounded perturbations while maintaining smooth gradients during optimization.

The optimization objective takes the form
\[
\mathcal{L}(x)=\pm p_i(x)+\lambda\,\mathrm{TV}(\delta),
\]
where $\mathrm{TV}$ denotes a Total Variation regularization term and $\lambda$ controls the strength of the smoothness constraint.

The Total Variation penalty is introduced to discourage high-frequency perturbations and isolated pixel-level noise. Rather than allowing arbitrary unconstrained modifications, the regularization biases the optimization toward spatially coherent transformations distributed across the image.

Optimization is performed directly through backpropagation on the input image using the CLIP encoder as a differentiable mapping. The procedure is conceptually related to the generation of adversarial perturbations in deep neural networks \cite{Intriguing2013}. However, the objective here is not to induce arbitrary classification errors, but rather to navigate the intrinsic geometric organization already present in the embedding space. Specifically, we are moving along the main 
component of the data manifold.

\begin{figure}[ht]
\centering
\begin{tabular}{ccc}
    \includegraphics[width=.25\textwidth]{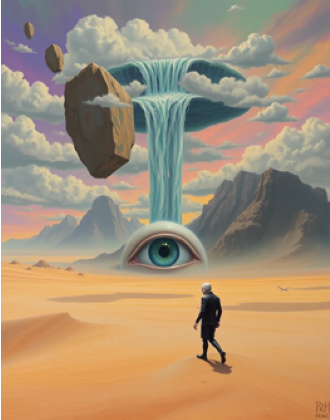}
    &
    \includegraphics[width=.25\textwidth]{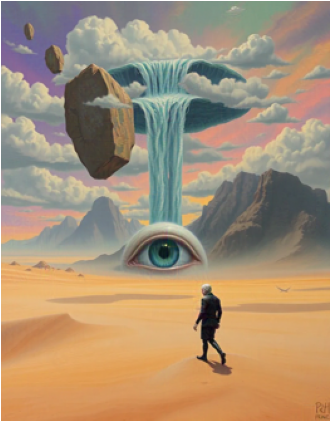}
    &
    \includegraphics[width=.25\textwidth]{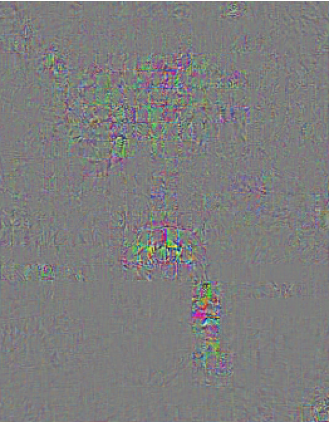}
    \\
    \includegraphics[width=.32\textwidth]{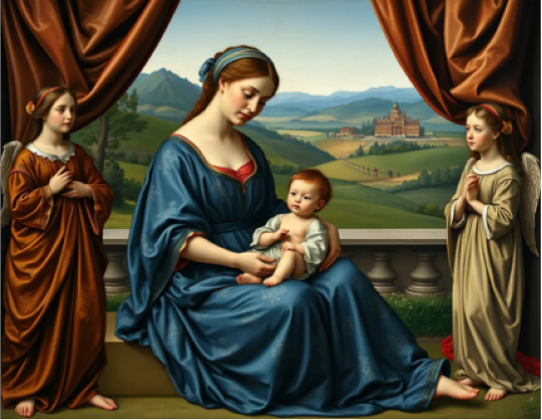}
    &
    \includegraphics[width=.32\textwidth]{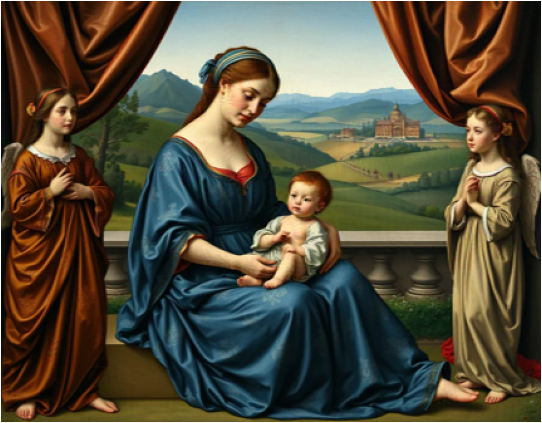}
    &
    \includegraphics[width=.32\textwidth]{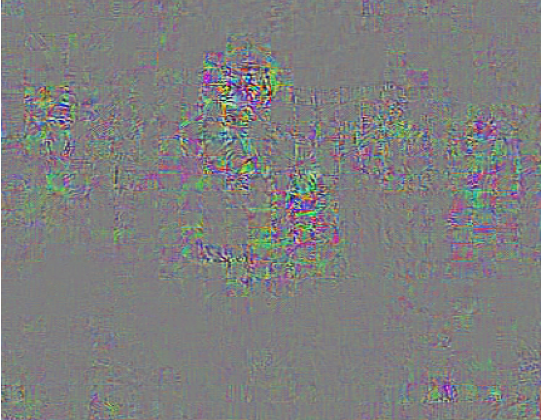}
    \\
    \includegraphics[width=.25\textwidth]{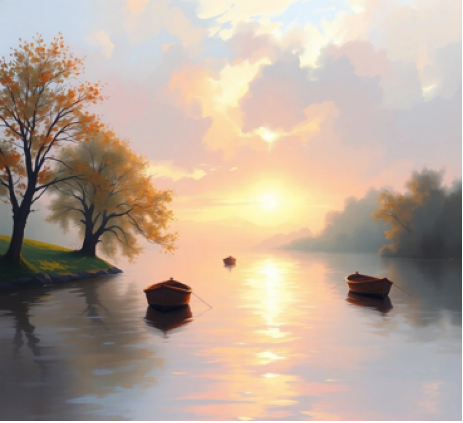}
    &
    \includegraphics[width=.25\textwidth]{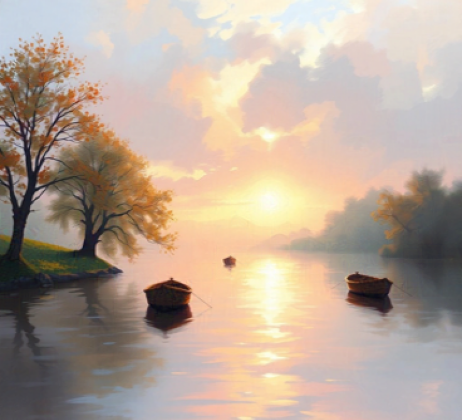}
    &
    \includegraphics[width=.25\textwidth]{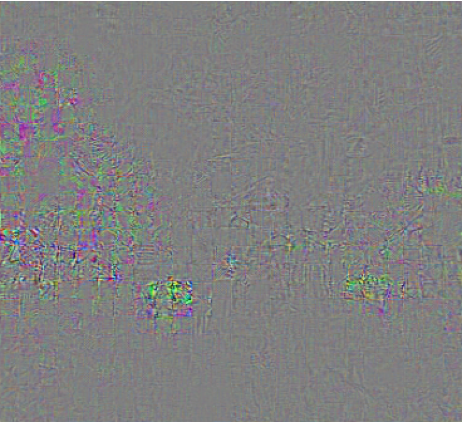}
    \\
    \includegraphics[width=.32\textwidth]{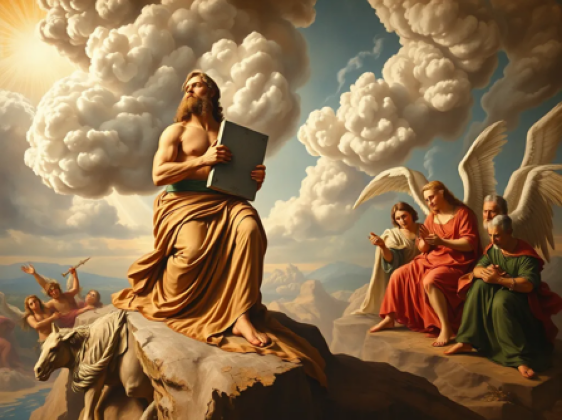}
    &
    \includegraphics[width=.32\textwidth]{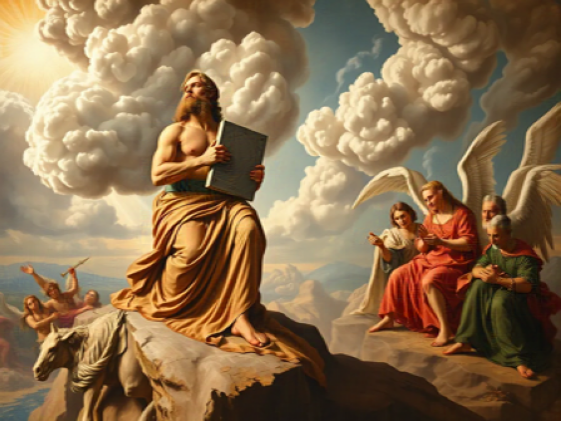}
    &
    \includegraphics[width=.32\textwidth]{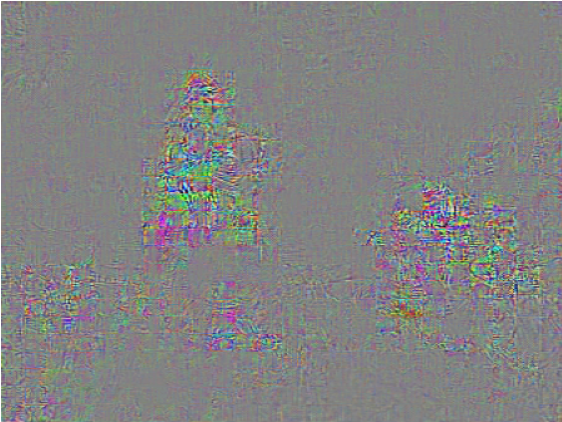}
    \\
    original & modified & difference
\end{tabular}
\caption{In the left column the original image, in the middle column the
modified version, and in the right column the difference between them.}
\label{fig:init_final_diff}
\end{figure}

Remarkably, substantial movements inside CLIP space can be obtained while the modified image remains visually almost indistinguishable from the original one, as 
exemplified in Figure~\ref{fig:init_final_diff}. 

\begin{figure}[ht]
    \centering
    \includegraphics[width=0.7\linewidth]{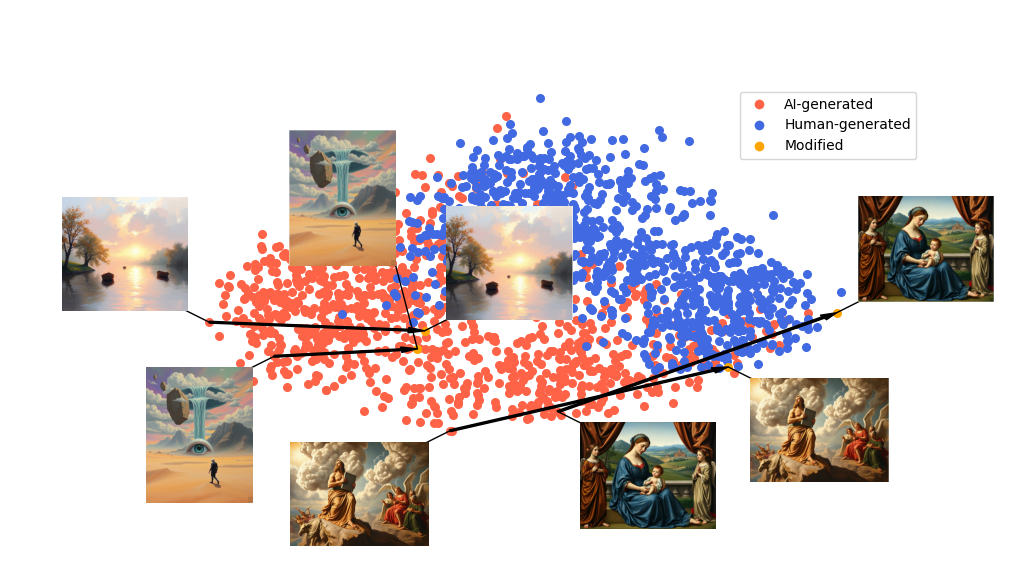}
    \caption{Movements in the 2D latent space induced by gradient ascent through CLIP along the main PCA component. Opposite movements for reproductions of human paintings can be easily generated too.}
    \label{fig:clip_inversion_movements}
\end{figure}

In particular, perturbation amplitudes remain numerically small (below 0.05 on the unitarian scale, see Table~\ref{tab:movements}) and visually coherent, while the corresponding displacement along the principal CLIP directions can become very significant, as
shown in Figure~\ref{fig:clip_inversion_movements}.

\begin{table}
    \centering
    \begin{tabular}{c|cccccc}
         image id & x init & y init & x final & y final & distance & max perturb.\\
         354 & -0.36 & -0.20 & -0.15 & -0.18 & 0.21 & 0.049 \\ 
         336 & 0.03 & -0.35 & 0.43 & -0.08 & 0.47 & 0.044 \\
         758 & -0.45 & -0.10 & -0.14 & -0.13 & 0.30 & 0.046 \\
         729 & -0.11 & -0.41 & 0.28 & -0.23 & 0.43 & 0.049 \\
    \end{tabular}
    \caption{Movements in the 2D PCA space induced by the gradient ascent along the main component. We provide initial and final coordinates, and euclidean distance. The perturbation is measured as  max absolute difference on images, normalized in the [0,1] interval.}
    \label{tab:movements}
\end{table}

Because gradient-based inversion through CLIP is computationally expensive, we limited the experiments to a representative subset of images. Nevertheless, all optimization experiments produced qualitatively similar results.

This phenomenon produces a genuine perceptual dissociation:
\begin{itemize}
\item from the human point of view, the optimized image remains essentially unchanged;
\item from the point of view of CLIP, the image undergoes a substantial transformation.
\end{itemize}

The importance of this result is not merely the existence of adversarial sensitivity, which is already well known in deep neural networks \cite{Intriguing2013}, but rather the fact that these perturbations operate along stable geometric directions that emerge spontaneously at the dataset level.

This suggests that the CLIP embedding space organizes visual similarity around structures that are only partially aligned with human perceptual salience. At the same time, an alternative interpretation remains possible: generative models themselves may implicitly introduce subtle statistical regularities that become highly coherent within the representation space while remaining nearly invisible to human observers.

The remainder of this work is devoted to investigating the nature of these hidden structures and identifying which statistical or perceptual properties may explain the observed separation.

\section{Methodological framework}
\label{sec:methodology}

Our investigation is structured into the following stages:

\begin{itemize}
\item Robustness analysis, aimed at evaluating the stability of the AI-human separation under a wide range of image transformations.

\item Basic statistical descriptors, investigating whether elementary image statistics related to color, luminance, and texture are sufficient to explain the dominant principal directions.

\item Advanced image descriptors, extending the analysis to progressively richer local and multiscale representations, with particular emphasis on Histograms of Oriented Gradients (HOG) and Scattering Transform features, while also investigating the role of their spatial organization.

\item Regression Model Inversion, validating the proposed statistical explanations by identifying image perturbations that induce controlled displacements within the PCA representation.
\end{itemize}

The four stages should not be regarded as independent experiments but as successive steps of the same investigation. Each stage is motivated by the limitations of the previous one and progressively narrows the range of plausible explanations for the observed separation. The following subsections describe each stage in greater detail.

\subsection{Robustness Analysis}
Before attempting to characterize the visual properties responsible for the separation between human and AI-generated paintings, we performed a set 
of preliminary experiments aimed ensuring that the phenomenon is genuine and not an artifact of a particular dataset, preprocessing pipeline, or model configuration. 

To this end, we performed a series of robustness experiments designed to assess the stability of the AI-human discrimination direction under a variety of transformations and experimental conditions.

The analysis considered file formats and preprocessing procedures, grayscale conversion and color manipulations, image resolution, and cropping strategies. 
Tests were performed on different AI generators, various human art collections, and multiple CLIP variants. Together, these experiments provide a systematic assessment of the stability of the observed separation and establish the empirical foundations for the subsequent analyzes.

\subsection{Basic Statistical Descriptors}

As a first step, we investigated whether the separation between human and AI-generated paintings could be explained by simple global image statistics. Such descriptors are attractive because they are easily interpretable and can often reveal systematic differences in color distribution, luminance, texture, or frequency content.

To this end, we considered several families of global descriptors, including pixel and luminance histograms, color statistics, radial frequency power spectra, and simple texture measures. We also examined correlations between these quantities and the principal directions of variation in the CLIP embedding space. The objective was to determine whether the AI-human separation could be attributed to a small number of low-level statistical properties before moving to more structured and spatially localized image representations.

\subsection{Advanced Image descriptors}

The limited explanatory power of simple image statistics, in combination with
the robustness of the AI-Human separation to major image perturbations like cropping, blurriness, or grayscale conversion, suggests that the signal could be better captured by more sophisticated statistics over larger image regions.

Motivated by the patch-based architecture of Vision Transformers, we investigated several families of patch-based representations.

Histogram of Oriented Gradients (HOG) \cite{dalal2005histograms} describes a patch through histograms of local gradient orientations and provides a compact representation of shape and texture. 

Scattering coefficients \cite{mallat2011group} provide a parametric family of
multi-scale and multi-orientation representations based on cascades of wavelet convolutions followed by modulus and averaging operations. Given a family of oriented wavelets, the parameter $J$ controls the maximum spatial scale represented in the decomposition, while a different parameter $L$ determines the number of sampled orientations. First-order coefficients capture the distribution of image energy across scales and orientations, whereas second-order coefficients additionally encode interactions between different scales and orientations. Unless otherwise specified, we employed second-order scattering coefficients and explored different values of $J$ and $L$ to assess the role of scale and angular resolution in the resulting representations.

In addition to HOG and Scattering, we also conducted experiments with 
Radial Frequency Power descriptors, which summarize the distribution of spectral energy across spatial frequencies by averaging the Fourier power spectrum over concentric frequency bands and Orientation Transport descriptors, which characterize the spatial organization and coherence of local orientations 
within the patch. However, these techniques didn't provide additional insight
into the problem, and they will not be discussed.

For HOG and Scattering, we devoted particular attention to determining whether the discriminative information arises primarily from the local image statistics themselves or from their spatial arrangement within the artwork. To investigate this question, we systematically evaluate several feature aggregation strategies for both descriptor families. These include patch-based feature extraction, which preserves local spatial information, as well as global aggregation methods based on both linear and nonlinear averaging (MLP, attention, single layer Transformers).

To further evaluate the role of spatial organization, we also performed patch-shuffling experiments in which patch descriptors were randomly permuted before being processed by the predictive model. Since shuffling preserves the distribution of local descriptors while destroying their spatial arrangement, the resulting performance provides a direct measure of the contribution of spatial structure to the discrimination task.

Together, these experiments allow us to distinguish between information encoded in the local statistics themselves and information arising from their organization across the image.

\subsection{Regression Model Inversion}

While the regression models provide a quantitative measure of how accurately the selected image descriptors predict the principal component coordinates, prediction accuracy alone is not sufficient to assess the quality or interpretability of a given descriptor. 
Empirically, many feature representations achieved excellent regression performance, yet failed to produce meaningful inversion results.
In other words, a low regression error does not necessarily imply that the learned mapping captures image characteristics that can be manipulated to produce a coherent displacement in the latent space. For this reason, regression performance should always be interpreted in conjunction with the corresponding inversion experiments, even though the two analyzes are presented separately in the following sections.

The main objective of the inversion process is to identify image modifications that induce a controlled displacement of an artwork within the two-dimensional PCA space. Starting from a given image, we seek perturbations that maximize or minimize the predicted PCA coordinates while preserving the perceptual appearance of the image. The inversion stage, therefore, constitutes the ultimate validation of a descriptor: a successful inversion demonstrates that the learned relationship between the descriptors and the PCA embedding is not only predictive but also actionable, enabling controlled navigation of the latent space through physically meaningful image modifications.

Throughout the optimization, perturbations are constrained to remain visually negligible, allowing the inversion process to reveal subtle statistical cues without altering the semantic content of the artwork.

Beyond visualization, inversion serves two complementary purposes. First, it discriminates between descriptors that merely correlate with the PCA coordinates and those that genuinely encode the visual information responsible for the observed separation. Only the latter produce stable and coherent image modifications capable of moving an image toward the region occupied by AI-generated or human-created artworks. Second, inversion provides an interpretable visualization of the image attributes that most strongly influence the learned representation, offering insights that cannot be obtained from regression accuracy alone.

%Rather than considering the PCA embedding as a static visualization, the inversion framework allows us to actively probe its geometry by analyzing how carefully controlled image perturbations translate into movement within the latent space. This perspective complements the quantitative regression analysis and provides a more comprehensive evaluation of the proposed descriptors, ultimately yielding a deeper understanding of the statistical characteristics that distinguish AI-generated images from human-created artworks.

\section{Robustness Experiments}
\label{sec:robustness}

\subsection{File format and preprocessing invariance}
\label{sec:preprocessing}

To assess whether the observed separation could be attributed to differences in file formats or preprocessing pipelines, we conducted a controlled re-encoding experiment. Each image was loaded, converted to a standard RGB representation, saved to disk in PNG format, and reloaded before being passed through the CLIP preprocessing pipeline. We then compared the resulting embeddings with those obtained from the original images using cosine similarity.

Across both human and AI-generated images, the cosine similarity between original and reprocessed embeddings was consistently equal to 1 (up to numerical precision). This indicates that the CLIP representation is effectively invariant to this re-encoding procedure, and that file format, compression artifacts, or metadata differences introduced at this stage do not contribute to the observed separability.

This experiment allows us to rule out a broad class of potential confounders, including differences in file format, compression schemes, metadata, and color encoding pipelines. Specifically, it suggests that the separation between human and AI-generated images is not driven by superficial artifacts introduced during image storage or loading.

\subsection{Color to Grayscale degradation}
\label{sec:grayscale}

Overall, as also testified by the statistics in 
Section \ref{sec:color palette}, 
color appears to play a more limited role in CLIP's visual representation than one might initially expect. To investigate this hypothesis, we compared the embeddings of the original images with those of their grayscale counterparts.

The average cosine similarity between the embedding of an image and that of its grayscale version is approximately $0.90 \pm 0.04$ for AI-Pastiche and $0.89 \pm 0.04$ for NGA. Thus, grayscale conversion produces only a relatively small displacement in CLIP's embedding space. 

\begin{figure}[ht]
    \centering
    \includegraphics[width=0.8\linewidth]{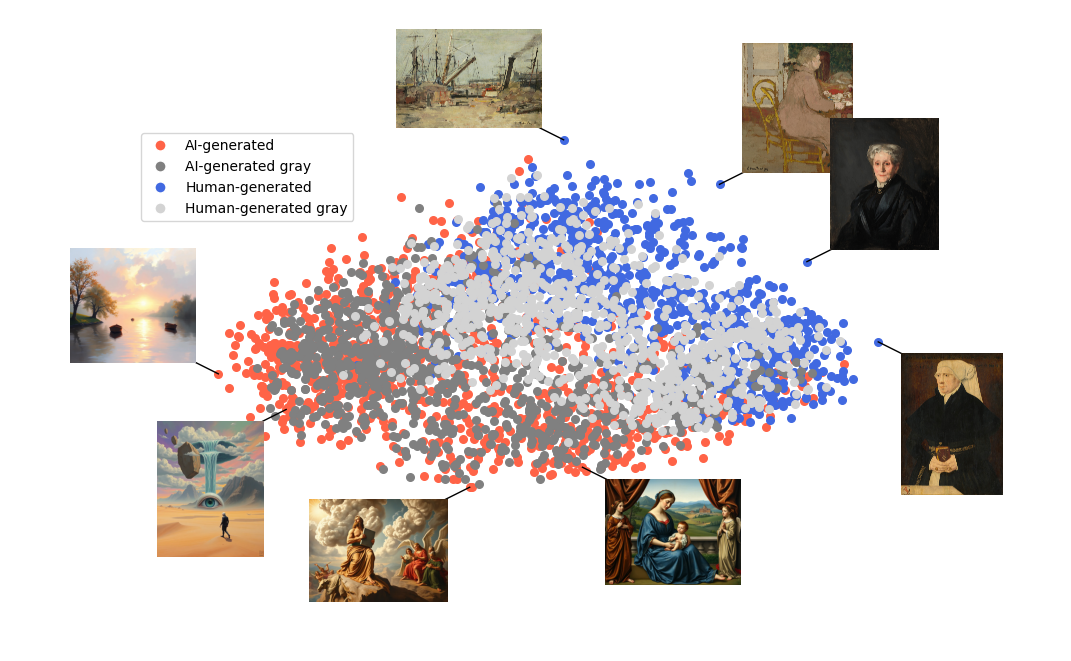}
    \caption{Projection of grayscale AI-Pastiche (dark gray) and NGA (light gray) images onto the PCA space computed from the original CLIP embeddings.}
    \label{fig:PCA_projection_gray}
\end{figure}

The distance becomes even smaller when projected into the 2D space.
Figure~\ref{fig:PCA_projection_gray} shows the projection of grayscale images onto the original PCA space defined by PC1 and PC2.

The two classes become slightly more entangled after grayscale conversion, suggesting that color contributes to the separation captured by PC1 and PC2. However, the effect remains limited, and most of the original structure is preserved. This indicates that the dominant differences between the two datasets are largely independent of color information.

\subsection{Resolution Robustness}

We next investigated the robustness of CLIP's representation to resolution degradation. The standard input resolution for the version of CLIP used in our experiments is $336\times336$ pixels. To evaluate the effect of image resolution, we first downsampled each image to a lower resolution and then resized it back to the standard input size before computing its embedding. We also performed analogous experiments using progressive Gaussian blurring and obtained qualitatively similar results.

Table~\ref{tab:resolution_similarity} reports the cosine similarity between the original embedding and the embedding obtained after the downsampling-upsampling process.

\begin{table}[]
    \centering
    \begin{tabular}{r|cccc}
        & $168\times168$ & $112\times112$ & $48\times48$ & $24\times24$ \\\hline
    AI-Pastiche & $0.93 \pm 0.02$ & $0.86 \pm 0.03$ & $0.72 \pm 0.06$ & $0.63 \pm 0.07$ \\
    NGA         & $0.93 \pm 0.02$ & $0.87 \pm 0.03$ & $0.67 \pm 0.06$ & $0.56 \pm 0.08$
    \end{tabular}
    \caption{Cosine similarity between the original CLIP embeddings and the embeddings obtained after resolution degradation.}
    \label{tab:resolution_similarity}
\end{table}

The embedding is only moderately affected by downsampling to $168\times168$ or $112\times112$ pixels. More severe degradation produces a larger displacement in the embedding space, although the images remain substantially correlated with their original representations.

\begin{figure}[ht]
\centering
\includegraphics[width=0.7\linewidth]{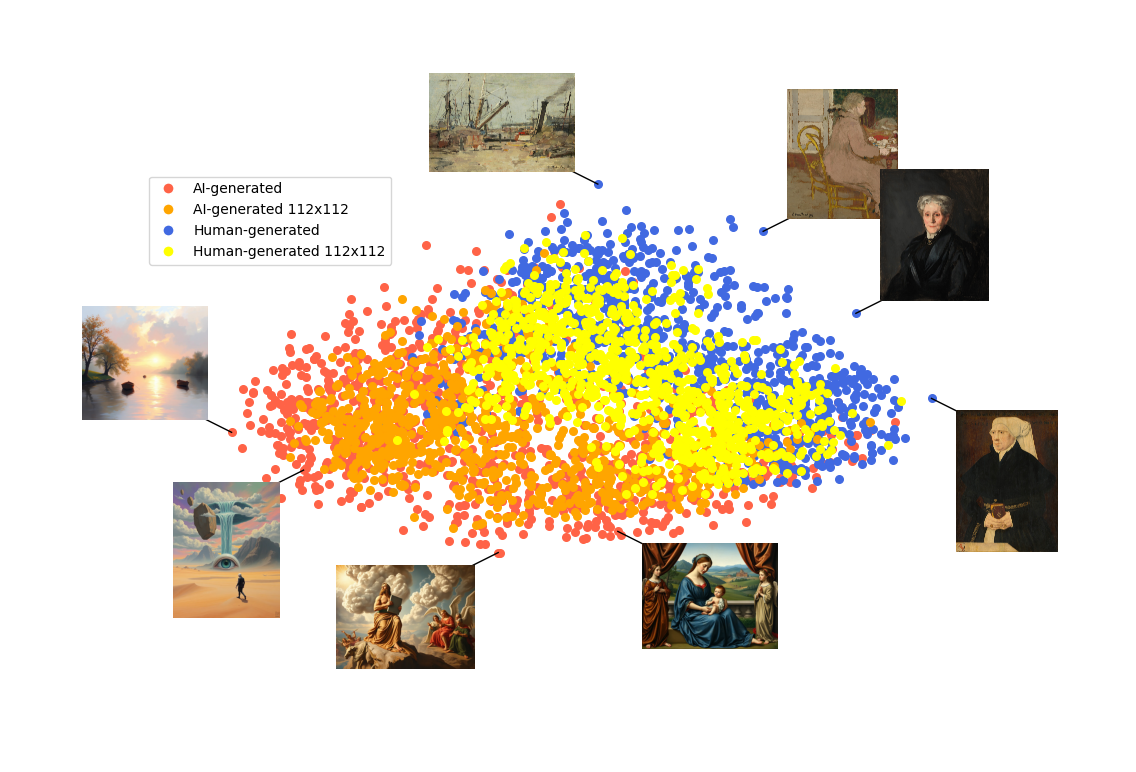}
\caption{Projection of images degraded to $112\times112$ pixels onto the original PCA space. Although the embeddings are displaced, the separation between AI-generated and human paintings remains largely preserved.}
\label{fig:PCA_112}
\end{figure}

\begin{figure}[ht]
\centering
\includegraphics[width=0.7\linewidth]{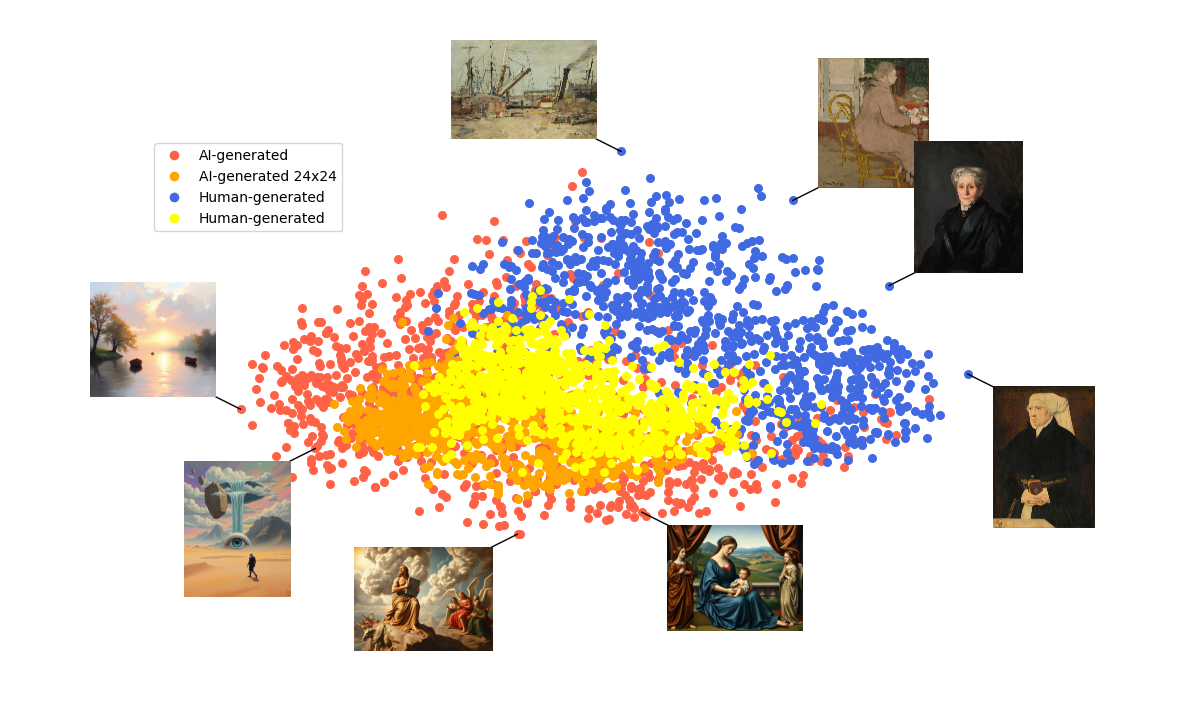}
\caption{Projection of images degraded to $48\times48$ pixels. The two distributions become more entangled, although a substantial fraction of AI-generated images remains separated from the human paintings.}
\label{fig:PCA_48}
\end{figure}

\begin{figure}[ht]
\centering
\includegraphics[width=0.7\linewidth]{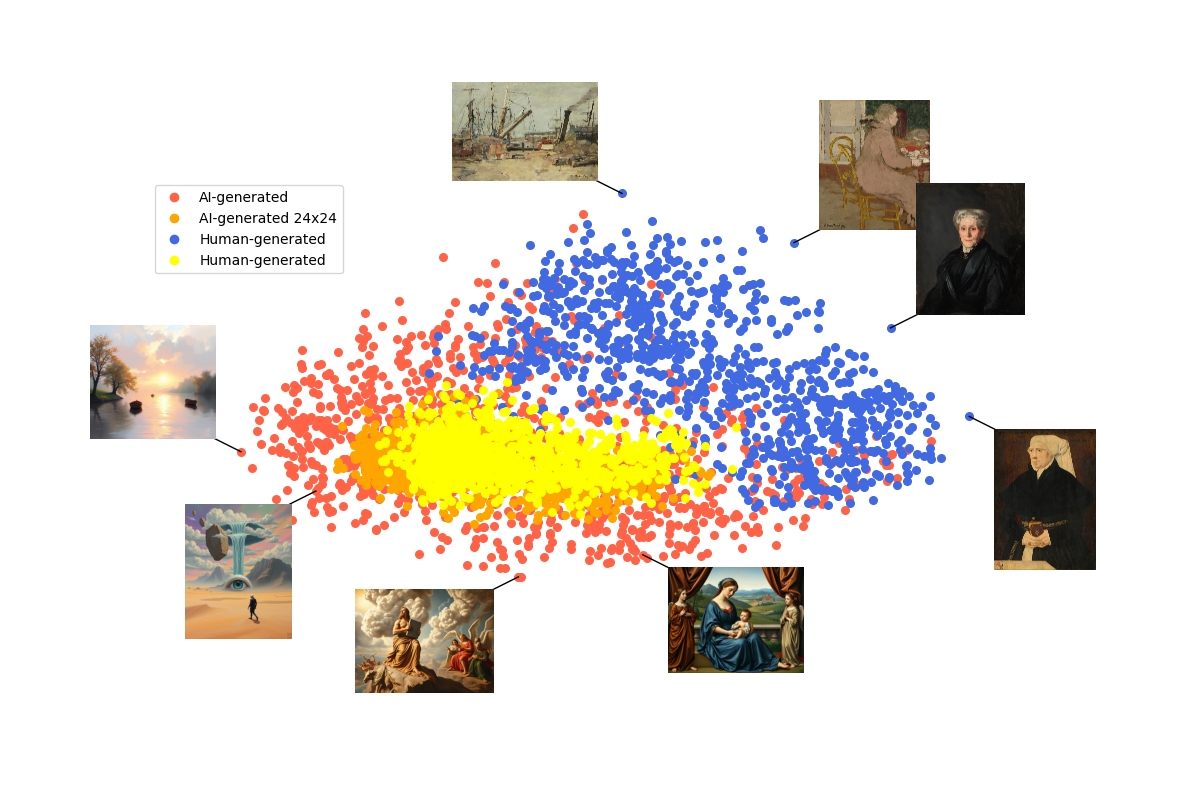}
\caption{Projection of images degraded to $24\times24$ pixels. At this resolution the two datasets become substantially entangled along the original PCA directions.}
\label{fig:PCA_24}
\end{figure}

The effect of resolution degradation can be visualized more clearly through the PCA projections shown in Figures~\ref{fig:PCA_112}, \ref{fig:PCA_48}, and \ref{fig:PCA_24}.

Overall, the separation between AI-generated and human paintings proves remarkably robust to substantial reductions in image resolution. This result suggests that the dominant CLIP components identified in our analysis are not driven by high-frequency details or fine-scale texture. Instead, they appear to reflect visual properties that remain stable across a broad range of spatial scales.

\subsection{Crop Robustness}

To distinguish between local and global sources of information, we investigated the effect of random image crops.

Table~\ref{tab:crop_classification} reports the performance of a logistic regression classifier trained to distinguish AI-generated from human paintings using CLIP embeddings computed from random crops of different sizes. Crop size is expressed as a percentage of the shortest image dimension. Consequently, a crop of $10\%$ corresponds to less than $1\%$ of the total image area.

\begin{table}[ht]
    \centering
    \begin{tabular}{ccccc}
        Crop size & Accuracy & Precision & Recall & F1 \\\hline
        $10\%$ & 0.516 & 0.516 & 0.515 & 0.515 \\
        $20\%$ & 0.763 & 0.767 & 0.763 & 0.762 \\
        $40\%$ & 0.943 & 0.946 & 0.945 & 0.945
    \end{tabular}
    \caption{AI vs. human classification performance using CLIP embeddings computed from random image crops. Classification is performed directly in the embedding space, before PCA projection.}
    \label{tab:crop_classification}
\end{table}

The results reveal a sharp transition as a function of crop size. Very small crops ($10\%$) contain essentially no discriminative information, whereas crops covering approximately $40\%$ of the shortest image dimension recover almost the full classification performance obtained using complete images.

Interestingly, crops of size $40\%$ remain strongly correlated with the embeddings of the corresponding full images. The average cosine similarity between crop and full-image embeddings is approximately $0.82 \pm 0.07$ for AI-Pastiche and $0.77 \pm 0.08$ for NGA.

\begin{figure}[ht]
\centering
\includegraphics[width=0.7\linewidth]{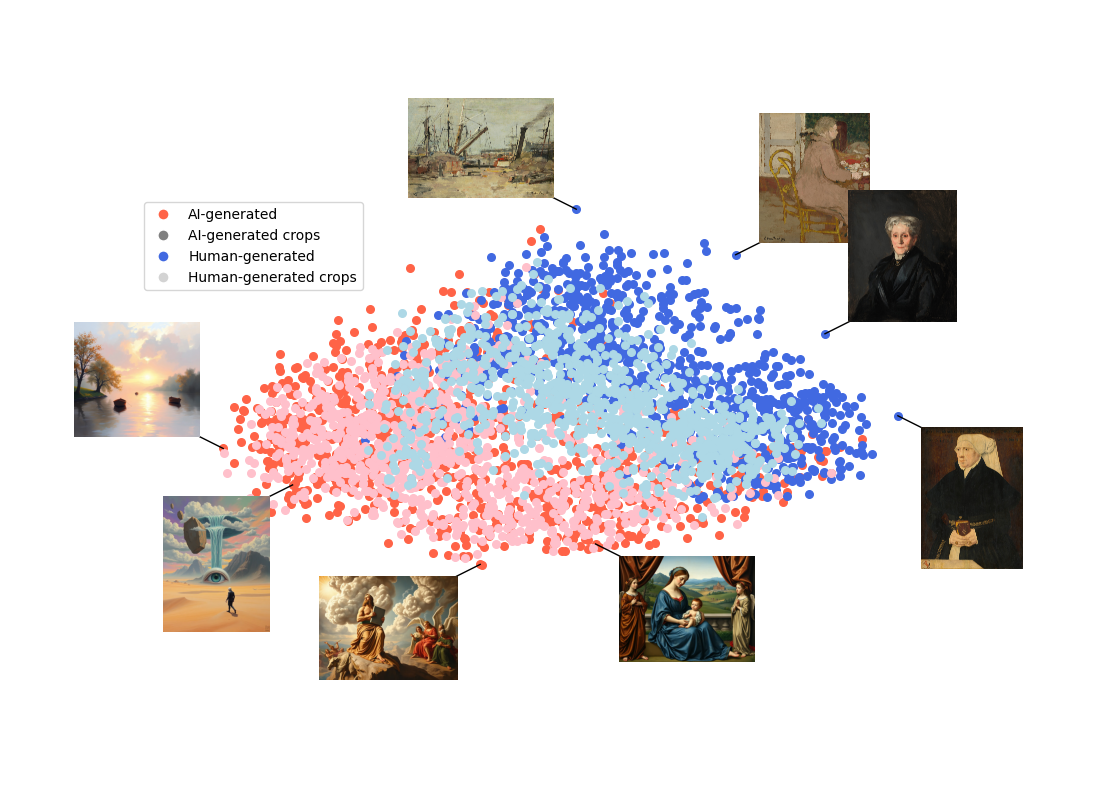}
\caption{Projection of random image crops onto the PCA space defined by the original CLIP embeddings. Although crops introduce additional variability, the separation between AI-generated and human paintings remains largely preserved for sufficiently large crop sizes.}
\label{fig:PCA_crops}
\end{figure}

Figure~\ref{fig:PCA_crops} shows the projection of cropped images onto the original PCA space.

These results allow us to rule out several simple explanations for the observed separation:

\begin{itemize}
%\item Watermarks or localized artifacts, which would remain detectable even in very small crops.
\item Fine-scale texture cues confined to small image regions.
\item Border or framing effects.
\item Purely global semantic or compositional structure, since crops covering only a fraction of the image remain highly discriminative.
\end{itemize}

Overall, the experiment suggests that the dominant discriminative signal is neither purely local nor purely global. Instead, it appears to emerge at an intermediate, mesoscopic scale, requiring sufficiently large image regions while remaining largely independent of the exact global image layout.

\subsection{Generator Robustness}

\begin{figure}[ht]
    \centering
    \includegraphics[width=0.9\linewidth]{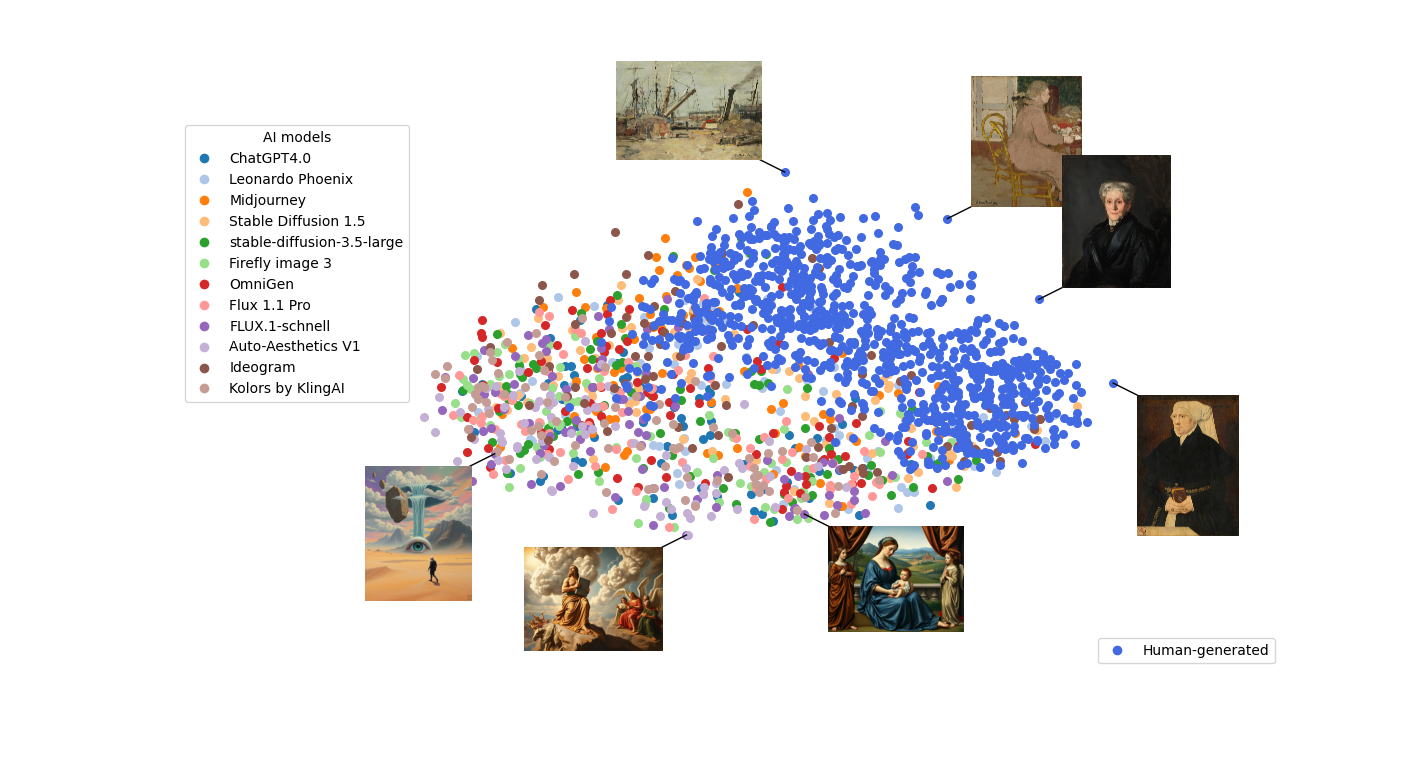}
    \caption{Distribution of AI-artworks for the different generative models 
    of AI-Pastiche. }
    \label{fig:all_models}
\end{figure}

The AI-Pastiche dataset covers a large collection of generative models belonging to different generations; hence, the separation problem does not seem to be reducible to a particular model or class. In Figure~\ref{fig:all_models} we show the disposition of the
AI-generated images for the different models in the first release of AI-Pastiche.

The distinction also persists in the most recent generators.
In Figure~\ref{fig:all_models} we plot the disposition of 
a very recent extension to the AI-Pastiche dataset, comprising Imagen4, FireflyImage4, Playgroundv2.5, Midjourney 7, Flux.2 and gpt-image-1.5.

\begin{figure}[ht]
    \centering
    \includegraphics[width=0.8\linewidth]{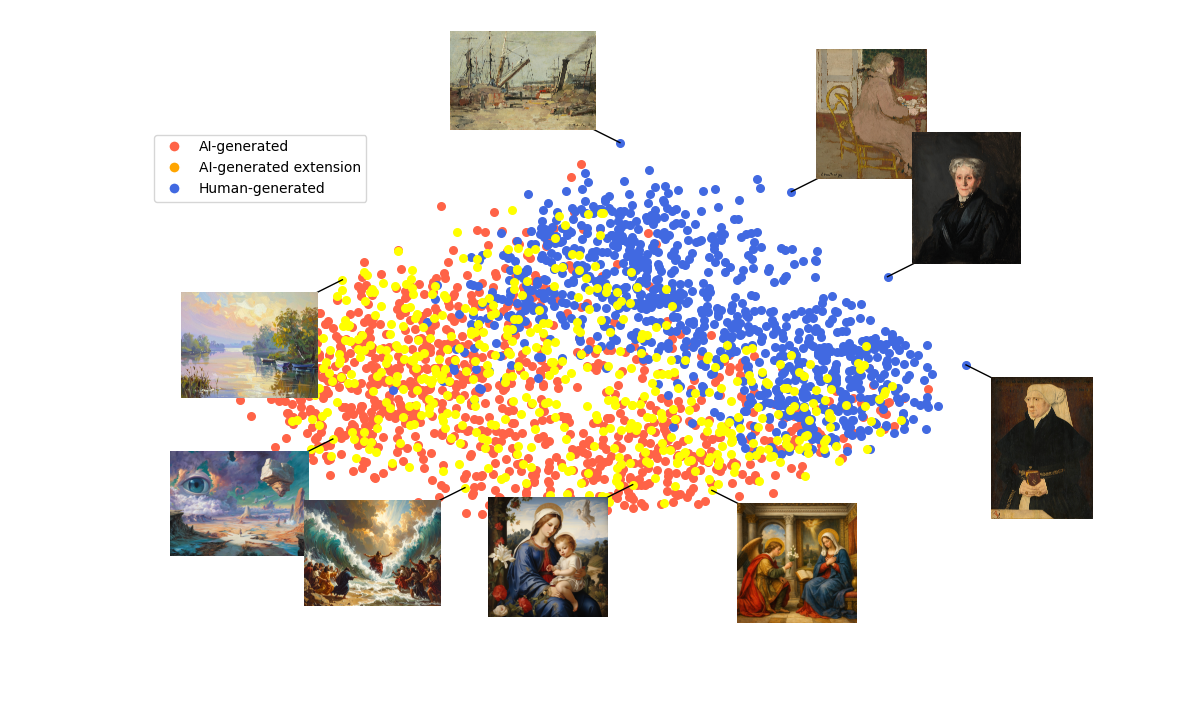}
    \caption{Distribution of AI-craftworks for generative models 
    of the latest generation. }
    \label{fig:new_models}
\end{figure}

Taken together, the robustness experiments indicate that the observed separation is remarkably stable across datasets, generators, preprocessing strategies, and image transformations. Consequently, the phenomenon is unlikely to be explained by simple image artifacts or isolated visual cues, motivating the search for increasingly expressive statistical descriptors in the following sections.

\section{Global Statistical Descriptors}
\label{sec:global_statistics}
Our investigation begins with the simplest explanatory hypothesis: that the separation between human and AI-generated paintings can be accounted for by elementary global image statistics. These descriptors are attractive because they are directly interpretable and have long been used to characterize color, luminance, texture, and frequency content. Their analysis also provides an important baseline, allowing us to determine whether the observed geometry of the CLIP embedding distribution can be explained by low-level statistical properties before introducing progressively richer image representations.

\subsection{Basic statistics}
Table~\ref{tab:basic_stats} reports the basic statistics for the two 
datasets along the three RGB components. Specifically, we 
give the min, max, mean, and (average) intra image standard deviation.

\begin{table}[ht]
    \centering
    \begin{tabular}{c|cccc|}
         %& \multicolumn{4}{c|}{AI-Pastiche} \\
        {\bf AI-Pastiche} & min & max & mean & std \\
        red & $5.7\pm9.3$ & $249.5\pm 9.2$ & $116.7 \pm 63.3$ & $63.4\pm13.2$ \\
        green & $6.6\pm9.9$ & $242.1\pm 12.9$ & $105.7 \pm 88.2$ & $57.4\pm 14.3$ \\
        blue& $4.0\pm8.1$ & $229.5\pm 21.7$ & $116.7 \pm 63.4$ & $54.4 \pm 14.3$\\\hline
        %& \multicolumn{4}{c|}{NGA}\\
        {\bf NGA}& min & max & mean & std \\
        red & $20.2\pm 26.5$ & $239.8\pm 18.3$ & $118.6\pm 43.1$ & $48.0\pm 15.6$ \\
        green & $20.9\pm 22.7$ & $224.0\pm 26.7$ & $104.1\pm 42.8$ &
        $42.9 \pm 14.6$\\
        blue & $13.2\pm 20.6$ & $200.7\pm 37.0$ & $82.2\pm 41.7$ &
        $39.7\pm 16.8$\\
    
    \end{tabular}
    \caption{Basic statistics about AI-Pastiche and NGA data.}
    \label{tab:basic_stats}
\end{table} 

Several differences are statistically apparent, particularly in the blue channel and in the intra-image variability. However, these differences are relatively modest compared with the magnitude of the separation observed in CLIP space and do not provide a convincing explanation for the dominant principal directions.

The following table reports the Shannon entropy of the empirical distribution of pixel intensities. For each image, all pixel values from the three color channels are pooled together and regarded as samples from a discrete distribution over the 256 possible intensity levels.

\begin{table}[ht]
    \centering
    \begin{tabular}{c|c|c}
                 &  {\bf AI-pastiche} & {\bf NGA}\\
         {\bf entropy} & $7.43 \pm 0.52$ & $7.05 \pm 0.63$
    \end{tabular}
    \caption{Entropy of the empirical distribution of pixel intensities for AI-pastiche and NGA.}
    \label{tab:placeholder}
\end{table}

\begin{figure}[ht]
\centering
    \begin{subfigure}[b]{0.49\textwidth}
    \centering
    \includegraphics[width=5.5cm]{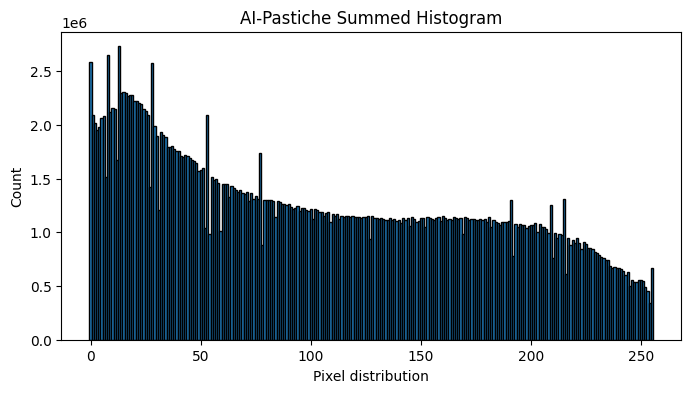}
    \caption{\label{fig:image1}}
    \end{subfigure}
    \begin{subfigure}[b]{0.49\textwidth}
    \centering
    \includegraphics[width=6.5cm]{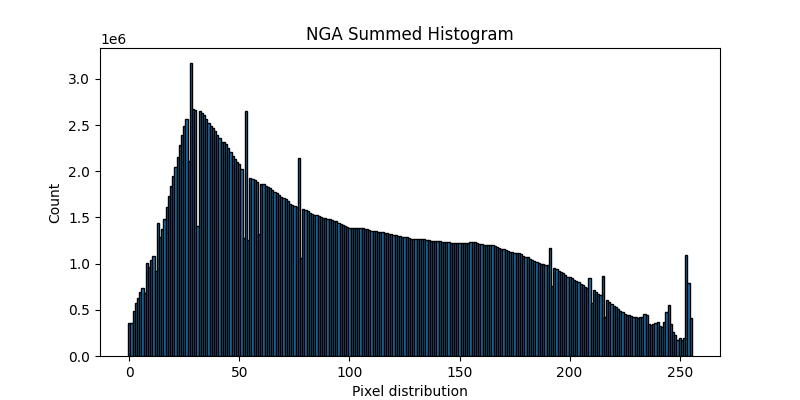}
    \caption{\label{fig:image2}}
    \end{subfigure}
    \caption{Histograms of distribution of pixel intensities for AI-pastiche and NGA.}
    \label{fig:histograms}
\end{figure}

In Figure~\ref{fig:histograms} we visualize the histogram of the distribution of pixel intensities for the two datasets. 

To assess whether the observed differences are primarily driven by global intensity statistics, we also evaluated several normalization procedures. None of them produced a substantial modification of the embedding geometry.

\subsection{Color Palette}
\label{sec:color palette}

\begin{figure}[ht]
    \centering
    \includegraphics[width=.8\linewidth]{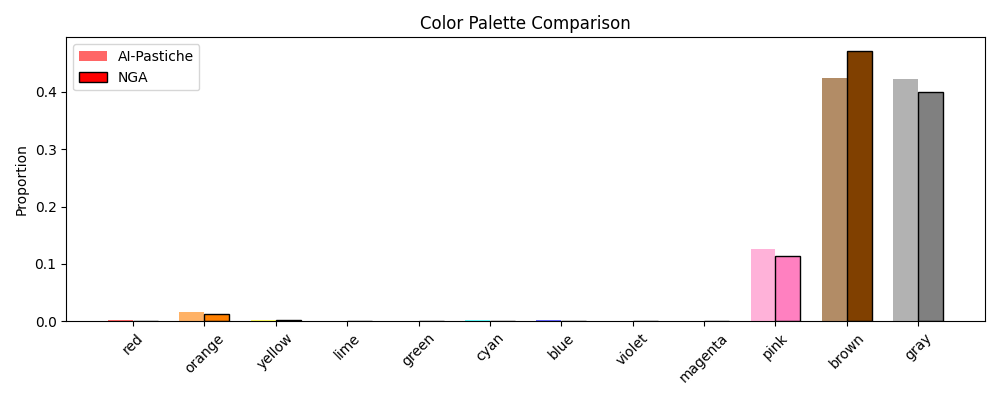}
    \caption{Empirical color palette distribution for the AI-Pastiche and NGA datasets.}
    \label{fig:color_palette}
\end{figure}

The color palettes of the two datasets are broadly similar, with a predominance of brown and gray tones (Figure~\ref{fig:color_palette}).

Numerical values are reported in Table~\ref{tab:color_palette}.

\begin{table}[ht]
    \centering
    \setlength{\tabcolsep}{3pt}
    \begin{tabular}{r|cccccccccccc}
              & red & orange & yellow & lime & green & cyan & blue
              & violet & magenta & pink & brown & gray \\
  {\bf AI} & .002 & .016 & .002 & .001 & .000 & .001 & .002 & .000 & .000 & .126 & .424 & .423 \\
  {\bf NGA}         & .001 & .012 & .002 & .000 & .000 & .000 & .000 & .000 & .000 & .113 & .472 & .400
    \end{tabular}
    \caption{Relative frequency of dominant color categories in the two datasets.}
    \label{tab:color_palette}
\end{table}

Overall, the color distributions of the two datasets are remarkably similar, exhibiting only minor quantitative differences. This observation is consistent with the robustness experiments of Section~\ref{sec:robustness}, where the separation remained largely unchanged after grayscale conversion. Taken together, these results suggest that global color composition is unlikely to constitute the primary statistical cue underlying the observed geometry of the CLIP embedding space.

\subsection{Local structure: luminance and texture}
The first descriptors exhibiting a non-negligible correlation with the principal components are those related to luminance variability and local contrast.

\begin{table}[ht]
\begin{tabular}{lcccccc}
Metric & AI-Pastiche & Human & AI-Pastiche & Human & AI-Pastiche & Human\\\hline
   & \multicolumn{2}{c}{$336\times 336$} & \multicolumn{2}{c}{$168\times 168$} & \multicolumn{2}{c}{$84\times 84$} \\
%Entropy & $7.44 \pm 0.52$ & $7.05 \pm 0.63$ \\
Chroma  & $19.7 \pm 9.8$ & $19.1 \pm 8.6$ & $19.4 \pm 9.6$ & $19.0\pm 8.6$  & $19.43 \pm 9.58$ & $18.87\pm 8.58$\\
Saturation HSL & $0.36 \pm 0.15$ & $0.32 \pm 0.13$ & $0.34 \pm 0.14$ &  $0.32 \pm 0.13$ & $0.34 \pm 0.14$ &  $0.31 \pm 0.13$ \\
Luminance mean & $44.7 \pm 14.7$ & $44.2 \pm 16.8$ & $44.7 \pm 14.7$ & $44.2 \pm 16.8$ & $44.7 \pm 14.7$ & $44.2 \pm 16.8$\\
Luminance std  & $22.6 \pm  4.3$ & $17.2 \pm 5.6$ & $22.0 \pm 4.3$ & $16.8 \pm  5.6$ & $21.3 \pm 4.2$ & $16.3 \pm  5.6$\\
local L contrast & $6.8 \pm 2.2$ & $4.6 \pm 1.8 $ & $9.8 \pm 2.7$ & $6.9 \pm 2.4$ & $9.84 \pm 2.7$ & $6.8 \pm 2.4$\\
edge L contrast & $10.8 \pm 3.8$ & $7.1 \pm 2.9$ & $15.7 \pm 4.9$ & $10.7 \pm 4.0$ & $15.7 \pm 4.9$ & $10.7 \pm 4.0$\\\hline
\end{tabular}
\caption{Basic luminance and color statistics at different scales.}
\label{tab:l-statistics}
\end{table}
 
\begin{figure}[ht]
    \begin{subfigure}[b]{0.49\textwidth}
    \centering
    \begin{tabular}{c|cc}
     metric & $R^2 PC1$ & $R^2 PC2$ \\
     saturation &  0.13 & -0.15 \\
     chroma & 0.12 & 0.02 \\
     entropy & 0.31 & -0.05 \\
     luminance mean & 0.12 & 0.23\\
     luminance std & 0.35 & -0.26 \\
     local contrast & 0.33 & -0.23 \\
     edge contrast & 0.36 & -020\\
    \end{tabular}\vspace*{.8cm}
    \caption{$R^2$ Correlation between local structure statistics versus PC1 and PC2.}
    \label{tab:pca_correlations}
    \end{subfigure}
    \begin{subfigure}[b]{0.42\textwidth}
    \centering
    \includegraphics[width=0.9\linewidth]{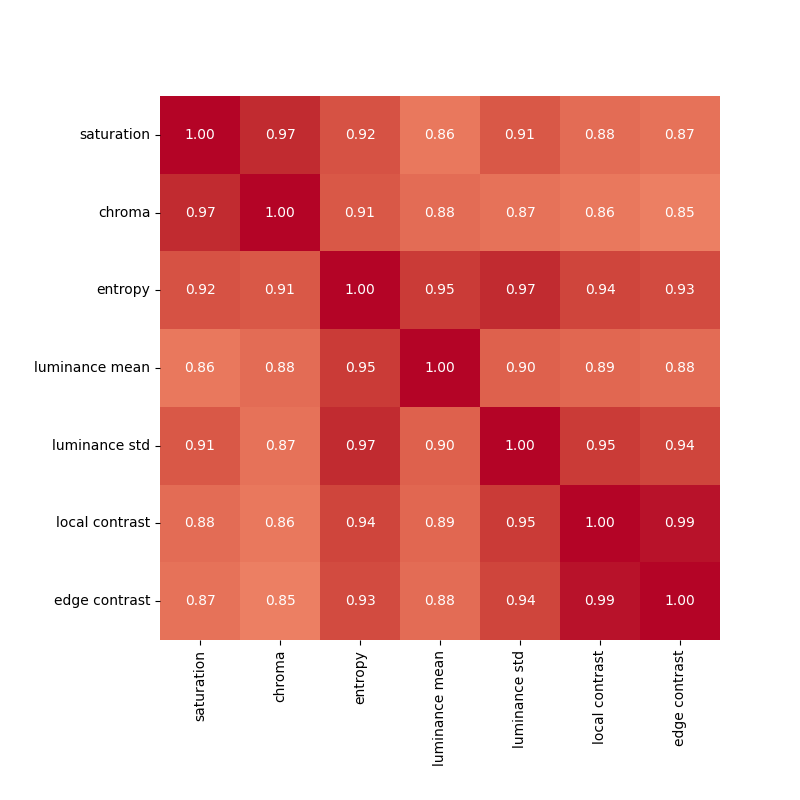}
    \caption{Self correlation between local structure statistics}
    \label{fig:features_correlations}
    \end{subfigure}
\end{figure}

Table~\ref{tab:pca_correlations} gives the correlation between the previous 
statistics with respect to the PC1 and PC2. 
A major problem is that all statistics are strongly correlated between each other (between 0.86 and 1), as reported in Figure~\ref{fig:features_correlations}. So, there is little to be gained from considering a linear combination of them. 

These observations suggest that simple global statistics and elementary measures of local contrast capture only a limited aspect of the phenomenon. This motivates the transition to descriptors capable of representing richer spatial organization, beginning with local gradient statistics in the following section.

\section{Histogram of Oriented Gradients}
\label{sec:HOG}
Histogram of Oriented Gradients (HOG) descriptors \cite{dalal2005histograms} characterize local image structure by accumulating gradient orientations into histograms computed over small spatial cells. In our implementation, each $24\times24$ patch was subdivided into a $2\times2$ grid of cells, each represented by a histogram of 9 orientation bins. Thus, every patch was encoded by a $2\times2\times9=36$ dimensional descriptor. Since a $336\times336$ image contains a regular grid of $23\times23$ overlapping patches (stride 14), the complete HOG representation forms a tensor of size $23\times23\times2\times2\times9$. Different aggregation strategies were subsequently investigated, ranging from global spatial averaging to preserving the complete descriptor field, as described in the next subsection. 

In Table~\ref{tab:hog aggregation statistics} we report
the $R^2$ similarity of different aggregations to PC1 and PC2.

\begin{table}[ht]
    \begin{tabular}{c|c|c|c}
         feature & size & PC1 $R^2$ & PC2 $R^2$\\\hline
         HOG field        & $23\times23\times2\times2\times9$ &0.30 & 0.32 \\
         field of HOG av. & $23\times23\times2\times2$ &0.24 & 0.25 \\
         field of HOG av. & $23\times 23$ & 0.10 & 0.11 \\
         HOG spatial av.  & 9 & 0.03 & 0.02 \\
         HOG spatial av.  & $2\times2\times9$ & 0.06 & 0.05
    \end{tabular}
    \caption{$R^2$ similarity of different aggregations of $24\times24$ patch based hog features to PC1 and PC2.}
    \label{tab:hog aggregation statistics}
\end{table}

The full field provides the best results, while the signal is nearly erased by spatial averaging.
%Interestingly, part of the HOG signal survives to spatial averaging, especially along PC2 that seems to describe more global signals.

\begin{figure}[htb]
        \begin{subfigure}[t]{0.32\textwidth}
            \centering
            \includegraphics[width=\linewidth,height=60mm, keepaspectratio]{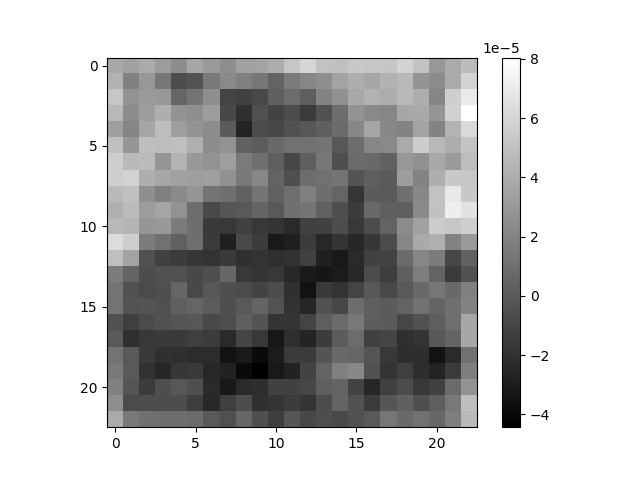}
            \caption{HOG mean}
            \label{fig:hog mean}
        \end{subfigure}
        \begin{subfigure}[t]{0.32\textwidth}
            \centering
            \includegraphics[width=\linewidth,height=60mm, keepaspectratio]{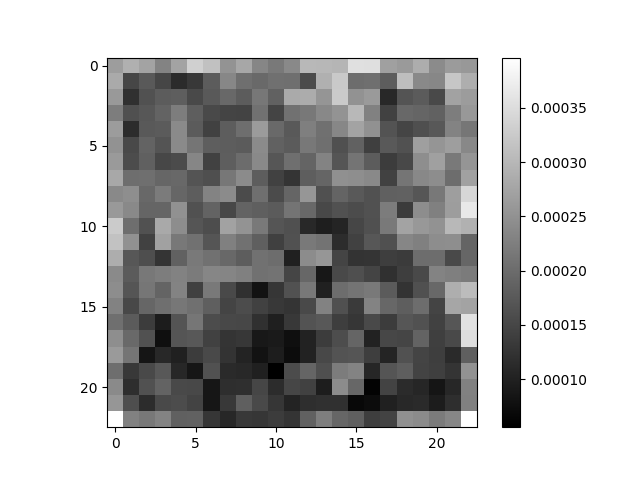}
            \caption{HOG max}
            \label{fig:hog mean}
        \end{subfigure} 
        \begin{subfigure}[t]{0.3\textwidth}
            \centering
            \includegraphics[width=\linewidth,height=60mm, keepaspectratio]{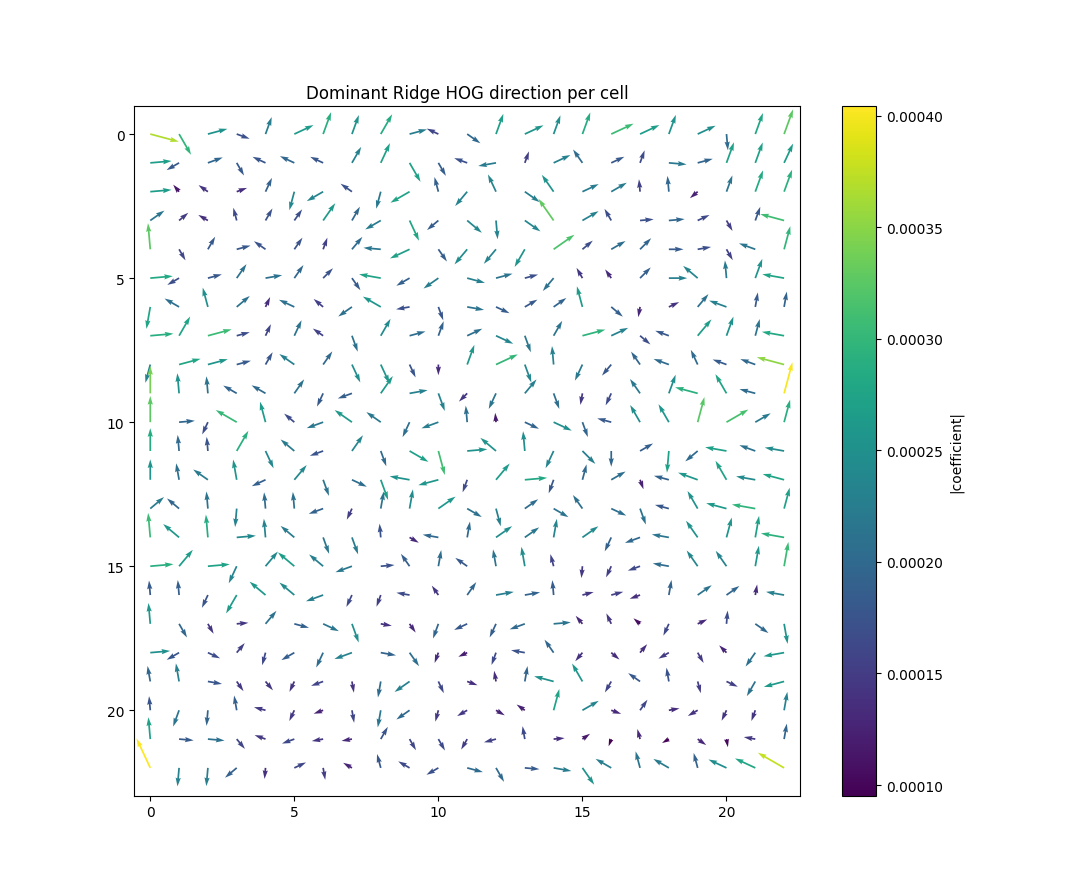}
            \caption{HOG orientations}
            \label{fig:hog orientations}
        \end{subfigure}  
        %\label{fig:hog visualizations}
        \caption{\label{fig:hog visualizations}Visualization of HOG statistics.}
\end{figure}

In Figure~\ref{fig:hog visualizations} we show the field of mean HOG values (a), max HOG values (b) and prevalent orientations(c).
The first two maps are relatively uniform (the range of values is very limited), while the third one is quite chaotic. 

A few observations can be drawn from the analysis. First, the discriminative signal does not exhibit sparse localization, indicating that it is not concentrated in a few isolated regions of the image. Likewise, there are no evident "artifact zones" that could be associated with localized generation defects. Instead, the observed signal has a relatively low amplitude but remains spatially coherent across the image. Overall, the resulting maps are smooth and broadly distributed, supporting the hypothesis that the distinction between AI-generated and human-created artworks is encoded in subtle, spatially organized statistical patterns rather than in isolated visual artifacts.

\subsection{Inter-patch investigations}
In the previous experiment, each patch provided an independent
linear contribution to the result. It is natural to wonder
if a better result can be obtained by comparing patches 
together, deriving non-linear dependencies. 

As a preliminary test in this direction, we investigated inter-patch features via non-linear methods, specifically a simple two layer MLP. 
As input features, we used 
the full HOG field. Due to the number of features, strong
regularization was required. We obtained the best results with $\alpha=20$ (the default is alpha=0.0001). All tests were obtained through Kfold cross validation with 5 splits. 
Results are provided in Table~\ref{tab:mlp_hog}.
In addition to MPL, we also tested Ridge over quadratic features (projected into 100 PCA components) but did not obtain significant results. 

\begin{table}[ht]
    \begin{tabular}{ccccc}
         model & features &  PC1 $R^2$& PC2 $R^2$ \\\hline
         MLP   & $23\times23\times2\times2\times9$ & 0.39 & 0.41         
    \end{tabular}
    \caption{$R^2$ similarity over a field of HOG features processed by a shallow MLP.}
    \label{tab:mlp_hog}
\end{table}

The previous result seems to confirm that spatial arrangement matters. At the same time. MLP only modestly improves over Ridge,
so the signal is not dominated by strong nonlinear interactions.

Overall, the results suggest that the distinction between AI-generated and human-created artworks is primarily encoded in weak but spatially distributed gradient statistics. Although the discriminative signal is subtle, it exhibits a consistent spatial organization rather than being confined to isolated image regions. Furthermore, the experimental results indicate that most of the relevant information can be effectively captured through linear aggregation of local descriptors, while more sophisticated nonlinear aggregation strategies provide only marginal improvements. This suggests that the underlying discriminative patterns are largely additive in nature, with nonlinear processing offering only limited refinement beyond the information already contained in the aggregated local statistics.

Next, we try to understand at what spatial scale the spatial arrangement becomes meaningful. We do it by shuffling HOG cells within local neighborhoods at progressive scales.

\begin{table}[ht]
    \centering
    \begin{tabular}{cccccc}
       block size & 1 & $2\times2$ & $4\times4$ & $8\times8$ & $23\time23$\\\hline
       PC1  & 0.39 & 0.36 & 0.30 & 0.27 & 0.04 \\
       PC2 &  0.41 & 0.37 & 0.29 & 0.22 & 0.16
    \end{tabular}
    \caption{HOG field shuffling, processed through MLP. The block size is the area being shuffled.}
    \label{tab:hog shuffling}
\end{table}

Table~\ref{tab:hog shuffling} shows that preserving the spatial organization of local HOG descriptors is essential for accurately predicting the PCA coordinates. While shuffling small regions only moderately affects performance, progressively larger shuffling blocks lead to a substantial reduction in the correlation with the principal components. 
These results indicate that preserving the spatial organization of local image statistics is essential for explaining the dominant embedding directions.

\subsection{Local Correlation descriptors}

The previous experiments suggested that the predictive signal captured by HOG descriptors is spatially distributed and cannot be reduced to isolated local features. This naturally led to the hypothesis that the relevant information might be encoded not in the marginal statistics of local descriptors, but rather in their spatial relationships and correlations. 
To this aim, we constructed several families of descriptors based on correlations or similarities between neighboring patches.

The experiments were performed on different underlying representations, including local luminance averages, local luminance standard deviations, raw image patches, and local HOG descriptors. In all cases, neighboring spatial cells were compared at fixed offsets (horizontal, vertical, diagonal, and distance-two neighbors), and the resulting similarity distributions were summarized through low-dimensional statistics such as mean, standard deviation, and quantiles.

For HOG-based descriptors, neighboring cells were represented by their full $36$-dimensional orientation histograms. For luminance and raw-patch representations, analogous correlation statistics were computed directly on pixel intensities or local patch vectors.

\begin{table}[ht]
\centering
\begin{tabular}{lccc}
\toprule
Covariance descriptor & Correlated representation & PC1 $R^2$ & PC2 $R^2$ \\
\midrule
Coarse HOG covariance & block-level $36$-D HOG descriptors & 0.07 & 0.04 \\
Local HOG neighbor corr. &  neighboring $36$-D HOG cells& 0.18 & 0.33 \\
Luminance mean neighbor corr. & neighboring scalar luminance cells & 0.04 & 0.02 \\
Luminance-std neighbor corr. & neighboring scalar contrast cells &  0.02 & 0.05 \\
Raw patch neighbor corr. & neighboring raw patch vectors &  0.01 & 0.01 \\
%\bottomrule

\end{tabular}
\caption{Regression performance obtained using pairwise spatial descriptors computed from different local image representations. Neighboring regions were compared through correlation or covariance statistics, while the coarse HOG descriptor was obtained by partitioning the image into a $4\times4$ grid and computing covariance statistics between the corresponding block-level HOG descriptors.}
\label{tab:covariance_descriptors}
\end{table}

Despite the variety of descriptors considered, all these approaches exhibited limited predictive power. The resulting regression scores remained substantially below those obtained using the full spatial HOG representation. Together with the shuffling experiments, these results suggest that the discriminative information depends on spatial organization, but that this organization cannot be adequately described by simple pairwise relationships between neighboring regions. This motivates the search for representations capable of capturing richer multiscale interactions, which is the subject of the next section.

\section{Scattering}
\label{sec:scattering}
The scattering transform is parameterized by two quantities. The parameter $J$ specifies the number of dyadic scales (equivalently, the maximum spatial scale represented), while $L$ denotes the number of wavelet orientations sampled at each scale. Throughout this work, we employed Kymatio scattering descriptors with maximum order two. Consequently, the representation is not limited to first-order wavelet responses, but also includes the zeroth-order low-pass coefficient and second-order coefficients describing interactions between first-order wavelet energies across different scales and orientations. The dimensionality of the resulting descriptor is

\[
N(J,L)=1+JL+\frac{J(J-1)}{2}L^2,
\]

where the three terms correspond, respectively, to the zeroth-, first-, and second-order coefficients.

We performed two complementary classes of experiments:

\begin{itemize}
\item Global scattering, aimed at assessing how much information is contained in conventional scattering descriptors and how the predictive performance depends on the maximum wavelet scale $J$.
\item Patch scattering, investigating whether the same descriptors become more informative when they are organized as a set of regional descriptors rather than being globally pooled into a single representation.
\end{itemize}

\subsection{Global Scattering}

We first evaluated the predictive power of conventional global scattering descriptors. 
Scattering coefficients were computed over the entire image, varying both the maximum wavelet scale $J$ and the angular resolution $L$.
The resulting feature vectors were used to predict the first two CLIP principal components through Ridge regression, with the regularization parameter selected by cross-validation.

Table~\ref{tab:scattering_ridge} summarizes the results. 

\begin{table}[ht]
    \centering
    \begin{tabular}{ccccccc}
    \toprule
    $J$ & Max scale $2^J$ &L & Features & PC1 $R^2$ & PC2 $R^2$ & $\alpha$ \\
    \midrule
    3 & 8  & 4 & 61 & 0.33 & 0.40 & 1 \\
    3 & 8  & 8 & 217 & 0.33 & 0.44 & 5 \\
    4 & 16 & 4 & 113 & 0.32 & 0.44  & 1 \\
    4 & 16 & 8 & 417 & 0.37 & 0.47 & 5 \\
    5 & 32 & 4 & 181 & 0.35 & 0.46 & 1 \\
    5 & 32 & 8 & 681 & 0.40 & 0.48 & 10 \\
    6 & 64 & 4 & 265 & 0.37 & 0.47 & 1\\
    6 & 64 & 8 & 1009 & 0.41 & 0.49 & 20\\
%    \bottomrule
    \end{tabular}
    \caption{Ridge regression from order-two scattering descriptors at different maximum scales 
    and different orientations.}
    \label{tab:scattering_ridge}
\end{table}

Increasing the maximum scale consistently enlarges the descriptor dimensionality, as explicitly reported in Table\ref{tab:scattering_ridge}. The predictive performance 
increases accordingly, but with a relatively modest improvement, passing from $R^2\approx0.33$ to
$R^2\approx0.41$ for PC1, and from $R^2\approx0.44$ to $R^2\approx0.49$ for PC2.  
These results indicate that scattering coefficients capture a non-negligible fraction of the information encoded by the CLIP components. However, increasing the receptive field of the scattering transform alone provides diminishing returns. In particular, descriptors including wavelet interactions up to a spatial scale of $32$ pixels ($J=5$) already capture almost all the predictive information obtainable from globally pooled scattering features, while doubling the maximum scale to $64$ pixels produces virtually no further improvement.

This saturation suggests that the main limitation is not the intrinsic scale of the scattering representation, but rather the use of a single global descriptor. This observation motivates the second series of experiments, where scattering coefficients are extracted independently from multiple image regions before being combined by a learnable model.

Similarly to the case of HOG, before addressing patches, we
conducted a few preliminary experiments with MLPs. 

A small multilayer perceptron substantially improved predictive performance, reaching approximately R
2
=0.60 for both PC1 and PC2. This result demonstrates that scattering coefficients contain considerable information about the PCA coordinates. However, as discussed in Section~\ref{sec:inversion}, predictive performance alone should not be interpreted as equivalent to explanatory power. In particular, the configurations achieving the highest regression scores did not produce the largest or most coherent displacements during inversion. The best inversion results were instead obtained for J=4 and L=4.

\begin{table}[ht]    
    \centering  
    \begin{tabular}{ccccccc}
    \toprule
    J & L & features & PC1 $R^2$ & PC2 $R^2$ \\
    \midrule
    3 & 4 & 61 & 0.54 &  0.57 \\
    3 & 8 & 217 & 0.55 & 0.59 \\
    4 & 4 & 113 & 0.56 & 0.60  \\
    4 & 8 & 417 & 0.58 & 0.60 \\
    5 & 4 & 181 & 0.58  & 0.61  \\
    5 & 8 & 681 & 0.60 & 0.61 \\
    6 & 4 & 265 & 0.60  & 0.60 \\
    6 & 8 & 1009 & 0.60 & 0.60 \\
    \bottomrule
    \end{tabular}
    \caption{Non linear (MLP) predictors from scattering descriptors. Results measured with Kfolds over 5 splits. Dropout 0.3, 200 epochs.}
    \label{tab:placeholder}
\end{table}

\subsection{Patch Scattering}
The previous experiments established that scattering descriptors contain substantial information about the dominant PCA coordinates.
While linear models achieved moderate predictive performance, a nonlinear MLP achieved $R^2$ values of approximately 0.60 for both principal components.
Since the descriptors were computed using order-two scattering transforms, these results already incorporate information about multiscale correlations among oriented structures, rather than merely local edge statistics. 

The objective of the patch-based analysis was, therefore, not simply to improve regression accuracy, but to determine whether explicitly preserving the spatial organization of scattering responses yields a representation more closely associated with the image variations underlying the PCA directions.
We experimented with several patch sizes and scattering configurations. For $84\times84$ patches, we considered $J=5$ with both $L=4$ (181 coefficients) and $L=8$ (681 coefficients). For $56\times56$ patches, we used $J=4$, $L=4$ (113 coefficients); for $28\times28$ patches $J=3$, $L=4$ (61 coefficients); and for $112\times112$ patches $J=5$, $L=4$ (181 coefficients). These configurations make it possible to investigate independently the influence of patch size, maximum represented scale, and angular resolution.

For each image, RGB values were converted to grayscale and partitioned into overlapping square patches. A second-order scattering transform was computed independently on each patch, producing one feature vector per patch. Depending on the values of $J$ and $L$, each patch was represented by 61, 113, 181, or 681 coefficients. The collection of patch descriptors forms a spatial feature map whose dimensions depend on the patch size (e.g., a $4\times4\times681$ tensor for $84\times 84$ patches). 
To avoid an explosion in the number of parameters, we mostly
worked with L=4. A few experiments with L=8 yielded only
marginal improvements.

Patch descriptors were processed by different lightweight models to assess
various aggregation techniques. 
Given the relatively strong performance of the linear model on global scattering statistics, we used its prediction as a baseline and trained the patch-based models to estimate a residual correction.
We tested a single-layer Transformer and a multi-head attention layer, both with four heads.
In addition, we tested both models in two different configurations,
one with an additional classification token (CL), and one without CL but with a self-learned positional encoding. We applied independent standardization over input Scattering coefficients, consistently with the preprocessing adopted by the Ridge baseline.

We also tested different models, comprising a few convolutional neural networks, without obtaining performance improvements.

\begin{table}[ht]
    \centering
    \begin{tabular}{c|c|c|c|c|c|c|c|c|c|c}
         \multicolumn{3}{c}{}& \multicolumn{4}{|c|}{$R^2$ PC1}& \multicolumn{4}{c}{$R^2$ PC2} \\
         \multicolumn{3}{c}{}& \multicolumn{2}{|c}{transformer} & \multicolumn{2}{|c}{attention} & \multicolumn{2}{|c}{transformer} & \multicolumn{2}{|c}{attention} \\
         patch dim & J & features & CL & pos & CL & pos & CL & pos & CL & pos\\\hline 
         %$14\times14$ & 2 & $25\times24\times 24$ & 0.45 & 0.46 & 0.40 & 0.44 & & & \\
         $28\times28$ & 3 & $61\times12\times12$ & 0.50 & 0.56 & 0.43 &  0.50 &  0.50 & 0.54 & 0.47 & 0.49 \\ 
         $56\times56$ & 4 & $113\times6\times6$ & 0.51 & 0.57 & 0.47 & 0.52 & 0.53 & 0.55 & 0.52 & 0.53 \\  
         $84\times84$ & 4 & $113\times4\times4$ & 0.51 & 0.57 & 0.50 & 0.52  & 0.51 & 0.54 & 0.48 & 0.52\\  
         $112\times112$  & 5 & $181\times3\times3$ & 0.52 & 0.58 & 0.52 & 0.58  & 0.51 & 0.53 & 0.52 & 0.52
    \end{tabular}
    \caption{PC1 and PC2 regression values for a lightweight model acting residually over a linear baseline. We tested a single layer Transform and a pure multi-head attention layer, both with four head. The two models have been tested in two configurations: with a classification token, and with positional encoding with averaging. }
    \label{tab:patch_scattering}
\end{table}

The resulting predictive performance generally lies between that of the linear and MLP models. More importantly, despite explicitly preserving the spatial organization of the scattering coefficients, the patch-based models did not produce a corresponding improvement in the inversion experiments. Thus, increasing the sophistication of the aggregation mechanism improves neither the interpretability nor the controllability of the learned relationship. This result further emphasizes that regression accuracy alone is insufficient to identify the image statistics genuinely associated with movement along the principal CLIP directions.

\section{Inversion}
\label{sec:inversion}

To investigate the visual meaning of the CLIP principal components, we optimized an input image to maximize or minimize the score predicted by our differentiable scattering surrogate. Since the scattering descriptors and the Transformer regressors are fully differentiable, gradients can be propagated back to the input image.

Optimization starts from the original image $x_0$. Instead of optimizing the image directly, we introduce an unconstrained latent variable $u$ and define the image perturbation as

\[
\delta=\varepsilon\tanh(u),
\]

which guarantees that every pixel displacement remains bounded by $\pm\varepsilon$. Furthermore, updates are restricted to visually meaningful regions by multiplying the perturbation by a fixed spatial mask $M$ extracted from the original image. The mask is obtained from a smoothed edge map and concentrates the optimization budget around textured regions and object boundaries, while largely preserving homogeneous areas. The optimized image is therefore

\[
x=\mathrm{clip}(x_0+M\odot\delta,0,1).
\]

The optimization objective is
\[
\mathcal{L}(x)=
\mathrm{direction}\cdot s(x)
+\lambda_{\mathrm{TV}}\mathrm{TV}(\delta)
+\lambda_{\mathrm{HF}}\mathrm{HF}(\delta),
\]
where $\mathrm{direction}\in\{-1,+1\}$ determines the desired direction
of displacement along the selected principal component, the total variation
term encourages spatial smoothness, and the high-frequency penalty suppresses
perturbations dominated by fine-scale Fourier components.
Images were optimized for 150 iterations using the Adam optimizer with an initial learning rate of 1e-02. 

In Table~\ref{tab:movements_scattering} we report the movements 
obtained in the PCA space after re-projecting the image modified through
the inversion technique.

Although the induced displacements are substantially smaller than those obtained by directly inverting CLIP, their consistent magnitude indicates that scattering captures part of the visual information associated with the dominant CLIP directions. 

\begin{table}[ht]
    \centering
    \begin{tabular}{c|c|c|c}
         \multicolumn{2}{r}{}& \multicolumn{2}{c}{PC1 displacement}\\
         features & model & mean & std \\\hline\hline
         CLIP embeddinggs & CLIP & 0.079 & 0.056 \\\hline
         \multirow{2}{*}{hog field} & ridge & 0.010 & 0.09 \\
           & mlp & 0.09 & 0.07 \\\hline
         \multirow{2}{*}{global scattering J3 L4} & ridge & 0.016 & 0.017 \\
            & mlp  & 0.020 & 0.021 \\\hline
         global scattering J3 L8 & ridge & 0.015 & 0.016 \\\hline
         \multirow{2}{*}{global scattering J4 L4} & ridge & 0.018  & 0.020  \\
                    & mlp & 0.019  & 0.023 \\\hline  
         global scattering J4 L8 & ridge & 0.017 & 0.018  \\\hline  
         \multirow{2}{*}{global scattering J5 L4} & ridge & 0.016 & 0.017 \\
          & mlp & 0.016 &  0.022 \\\hline
         global scattering J5 l8 & ridge & 0.015 & 0.016 \\\hline
         \multirow{2}{*}{global scattering J6 L4} & ridge & 0.014 & 0.015 \\
                    & mlp & 0.015 & 0.023 \\\hline
         global scattering J6 L8 & mlp & 0.014 & 0.015 \\\hline\hline
         patch-28 scattering & Res. Attention & 0.017 & 0.019 \\\hline
         patch-56 scattering & Res. Attention & 0.018 & 0.020\\\hline   %new
         patch-84 scattering & Res. Attention & 0.018 & 0.019 \\\hline
         patch-128 scattering & Res. Attention & 0.017 & 0.019 \\\hline
         \multirow{2}{*}{global scattering J3-5} & Ridge Ensemble & 0.019 & 0.017 \\
                                                 & mlp Ensemble & 0.020 & 0.019 \\\hline
    \end{tabular}
    \caption{Summary of movements in the 2D PCA space induced by inversion of different regression techniques over different extracted features. The mean is computed over all images in AI-Pastiche.}
    \label{tab:movements_scattering}
\end{table}
The magnitude of the displacement obtained through scattering inversion is approximately one quarter of that obtained by direct CLIP inversion. While these quantities should not be interpreted as a direct measure of explained variance, the comparison provides a useful indication of the extent to which scattering captures the image variations to which CLIP is sensitive.

Compared with Ridge regression, the MLP provides a modest improvement in
the mean displacement for $J=3$ and $J=4$. This advantage disappears for
larger values of $J$: at $J=5$ and $J=6$, the mean displacement decreases
while its variance remains approximately unchanged. This suggests that the
nonlinear model primarily amplifies the inversion signal already captured
by the simpler representation. At larger scattering scales, the additional
information does not translate into a more coherent displacement, and the
variability across images becomes comparable to, or larger than, the mean
effect. In this regime, the inversion is therefore increasingly dominated
by image-dependent variability rather than by a stable direction associated
with the target principal component.

In Table\ref{tab:movements_scattering}, we only report the values for the Attention layers with self-learned positional encoding, but the results are similar for the other architectures. Explicitly preserving the spatial organization of patch-level scattering coefficients also provides no systematic improvement over the global scattering baseline. Thus, the information missing from the scattering representation does not appear to be recoverable simply by retaining the spatial arrangement of local scattering responses.

\begin{figure}[ht]
\centering
    \begin{subfigure}[b]{0.5\textwidth}
    \centering
    \includegraphics[width=7cm]{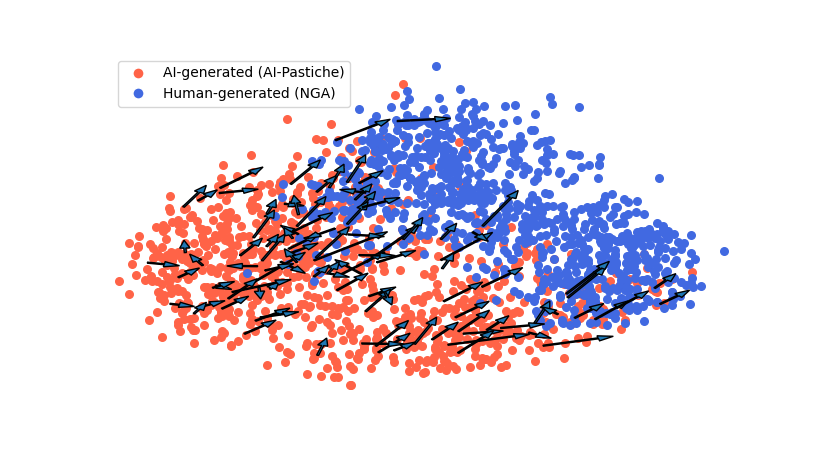}
    \caption{\label{fig:AI2Human}AI to Human}
    \end{subfigure}
    \begin{subfigure}[b]{0.45\textwidth}
    \centering
    \includegraphics[width=6.5cm]{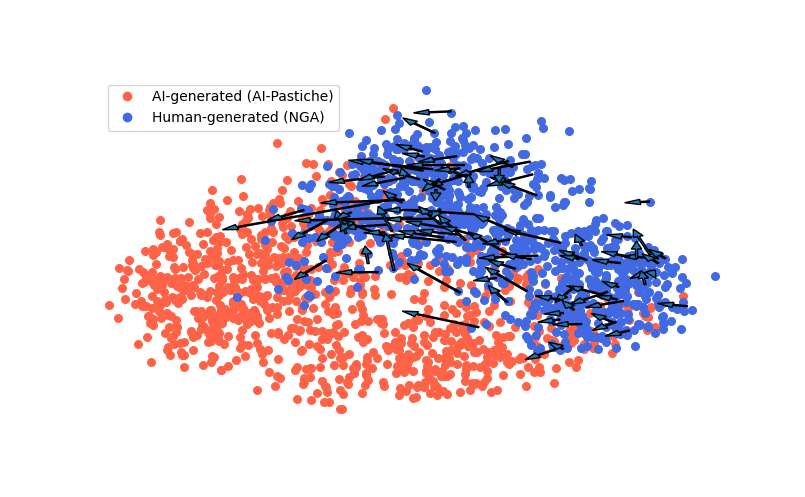}
    \caption{\label{fig:Human2AI}Human to AI}
    \end{subfigure}
    \caption{Movements in 2D space induced through scattering inversion. Movements are governed by the x-axis (PC1). }
    \label{fig:global_movements}
\end{figure}

In Figure~\ref{fig:AI2Human}, we show the effective displacement of one hundred random samples of AI-Pastiche towards the Human region. 

Similarly, in Figure~\ref{fig:Human2AI} we show the effective displacement of one hundred random samples of the paintings of the National Gallery towards the AI region. 

An important aspect of these results is the large variance of the induced displacement, also visible in the irregular trajectories of Figure~\ref{fig:global_movements}. While scattering inversion produces a systematic displacement on average, its effect varies considerably across individual images. This variability provides further evidence that scattering captures only part of the statistical information governing the CLIP principal directions.

In order to reduce the variance, we computed the inversion score
using an ensemble of the four ridge regressors for J=3 to J=6. 
Denoting by $s_j$ the prediction of the model associated with scattering features at J=j, the optimization objective is based on the ensemble score

\[
s(x)=\frac{1}{4}\left(s_3(x)+s_4(x)+s_5(x)+s_6(x)\right).
\]

This is the model in the last row of Table~\ref{tab:movements_scattering}. 
The motivation for the ensemble is that each model captures similar, but not identical, aspects of the CLIP components. During optimization, the four predictors therefore act as implicit regularizers for one another, discouraging image modifications that improve only a single surrogate while leaving the others unchanged. Working with the ensemble also
reduces the introduction of artifacts so typical of inversion techniques.
%and potentially allows for iteration over a larger number of steps.

\section{Discussion}
\label{sec:discussion}

\subsection{Interpreting the Scattering Representation}

Although the scattering transform provides a rich and stable representation of image textures, interpreting its coefficients is not straightforward. In particular, each coefficient is associated with a specific combination of scattering order, spatial scale, and orientation, making it difficult to directly relate individual coefficients to meaningful image characteristics. Moreover, the relatively large number of coefficients produced by the transform further complicates their visual inspection and comparison across different image datasets. Displaying the distribution of every coefficient separately would require dozens of plots, making it challenging to identify the dominant trends and differences.

To facilitate the interpretation of the scattering representation, we aggregate coefficients that share the same physical meaning. The objective is not to reduce the descriptive power of the scattering transform, but rather to obtain a compact representation that preserves its principal scale-dependent information while enabling an effective visual comparison between datasets.

For a scattering configuration adopted in this work, with $J=3$ scales and $L=4$ orientations, each image is represented by a vector of $61$ coefficients. These naturally decompose as
\[
61 = 1 + 12 + 48,
\]
corresponding to one zeroth-order coefficient, twelve first-order coefficients, and forty-eight second-order coefficients.

The first-order coefficients are organized into three spatial scales, with four orientations associated with each scale. Since the objective of this analysis is to compare the response at different scales rather than along individual directions, the four orientation responses are averaged, producing one coefficient for each scale. Consequently, the twelve first-order coefficients are reduced to three aggregated features.

The second-order coefficients describe interactions between pairs of scales. For $J=3$, only three scale combinations are possible,
\[
(j_1,j_2)\in\{(0,1),(0,2),(1,2)\},
\]
each containing all $4\times4$ orientation combinations. Averaging over these orientation pairs yields one coefficient for each scale pair, reducing the forty-eight second-order coefficients to three aggregated features.

The resulting representation, therefore, consists of only seven features:
\begin{itemize}
    \item $S_0$, the zeroth-order coefficient, representing the low-frequency average intensity;
    \item $S_1(0)$, $S_1(1)$, and $S_1(2)$, the first-order coefficients averaged over orientations, characterizing the image response at increasing spatial scales;
    \item $S_2(0,1)$, $S_2(0,2)$, and $S_2(1,2)$, the second-order coefficients averaged over orientation pairs, describing the interactions between structures occurring at different scales.
\end{itemize}

This aggregation reduces the dimensionality from $61$ to $7$ features while preserving the information associated with the different spatial scales. The resulting representation is considerably easier to interpret visually, allowing the distributions of the aggregated coefficients to be compared directly using histograms or other statistical visualizations.

\begin{figure}[ht]
    \centering
    \includegraphics[width=0.6\linewidth]{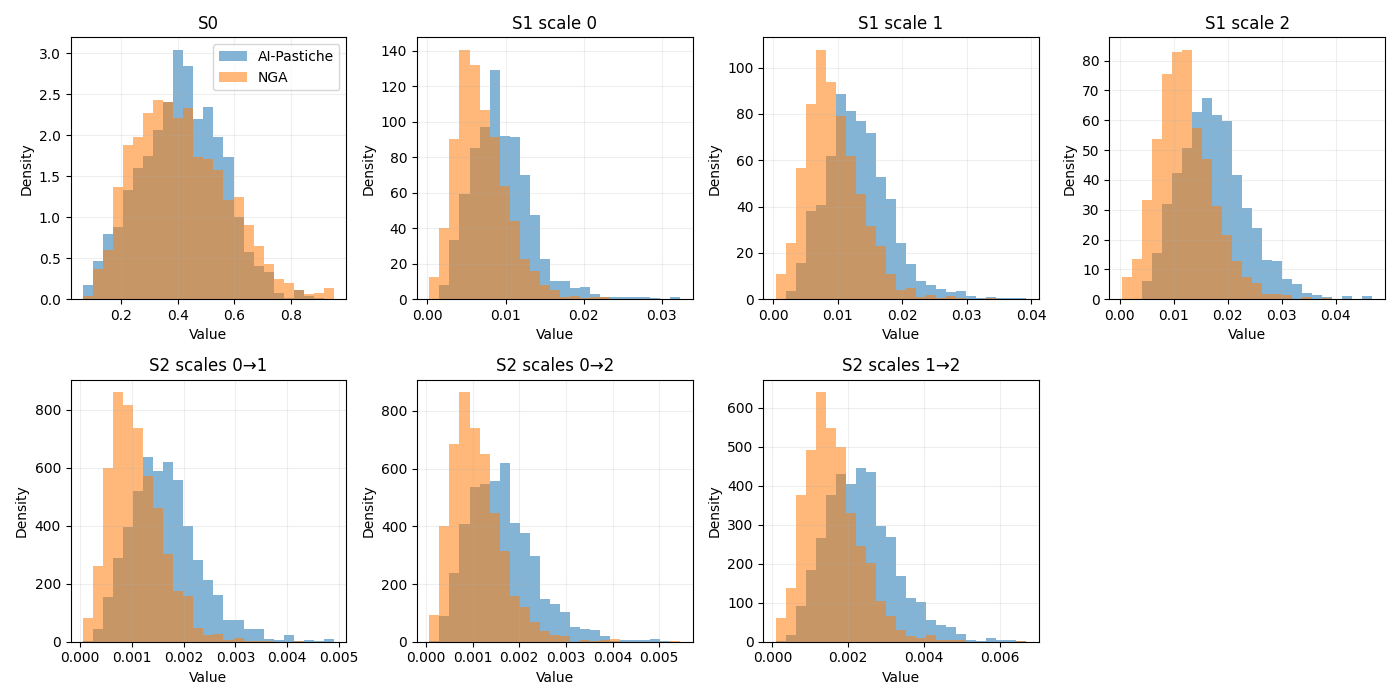}
    \caption{Distribution of scattering coefficients along aggregated groups at J=3 for works of AI-Pastiche (AI-generated) and NGA (human).}
    \label{fig:scattering_comparison_J3}
\end{figure}

Figure~\ref{fig:scattering_comparison_J3} describes the distribution of scattering coefficients at J=3 in the case of AI-Pastiche and paintings of the National Gallery of Art. 

\begin{figure}[ht]
    \centering
    \includegraphics[width=0.6\linewidth]{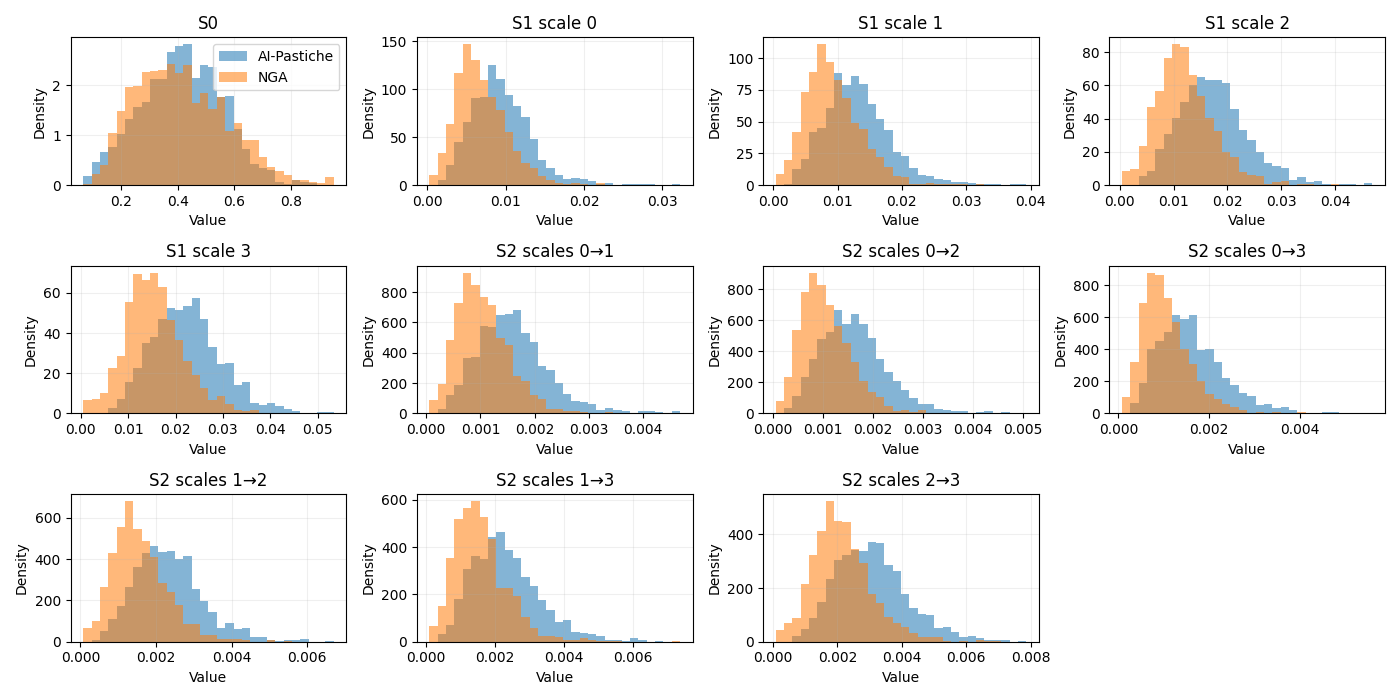}
    \caption{Distribution of scattering coefficients along aggregated groups at J=4 for works of AI-Pastiche (AI-generated) and NGA (human).}
    \label{fig:scattering_comparison_J4}
\end{figure}

We can repeat a similar computation at J=4, resulting in this case in 
11 resulting features, whose distribution is visualized in Figure~\ref{fig:scattering_comparison_J4}.

Despite the different scattering configurations, both figures exhibit the same overall behavior, indicating that the observed differences between AI-generated and human images are robust with respect to the choice of the maximum scattering scale.

The zeroth-order coefficient, $S_0$, shows a substantial overlap between the two datasets for both values of $J$, that is consistent 
with our previous discoveries.

A markedly different behavior is observed for the first-order coefficients. For every spatial scale, the distributions corresponding to the AI-generated images are consistently shifted towards larger values with respect to the NGA dataset. This indicates that the wavelet responses of the synthetic images are, on average, stronger across all analyzed scales. Moreover, the magnitude of the shift remains remarkably consistent as the scale increases, suggesting that this is a global characteristic of the generated images rather than an effect confined to a particular frequency range.

The most pronounced differences are observed for the second-order coefficients. For all combinations of scales, the distributions associated with the AI images exhibit larger values and a longer right-hand tail than those corresponding to the human images. Since second-order scattering coefficients quantify the interactions between image structures occurring at different spatial scales, these results suggest that AI-generated images possess stronger multiscale dependencies and more structured cross-scale correlations.

The analysis performed with $J=4$ further reinforces this interpretation. Besides reproducing the same behavior observed for the common features already present with $J=3$, the additional coarse-scale coefficients, namely $S_1(3)$, $S_2(0,3)$, $S_2(1,3)$, and $S_2(2,3)$, display the same systematic shift towards larger values for the AI dataset. Therefore, the observed discrepancy is not limited to fine or medium spatial scales, but also persists at coarser resolutions. This indicates that the differences between AI-generated and human images extend across the entire hierarchy of spatial scales captured by the scattering transform.

Considering scattering over spatial patches, it is possible to
observe that the spatial coefficient of variation $\sigma/\mu$
is approximately the same for the two sets of images. So
the larger scattering coefficients are not due to a regular,
repeated texture, but to an amplified scattering response: 
since the local scattering coefficients are globally larger, their absolute variance also increases, and the increase in variability is approximately proportional to the increase in the mean. 

\subsection{A Speculative Interpretation}
The previous analyses establish several robust empirical observations but do not by themselves explain the mechanisms responsible for them. In the following, we discuss two speculative hypotheses suggested by our experimental findings. These hypotheses are intended as possible interpretations rather than definitive explanations, and should therefore be regarded as directions for future investigation.

\subsubsection{Why do AI-generated paintings exhibit systematically larger scattering responses?} 
A possible explanation for the systematically larger scattering responses
observed in generated images is that they arise as a consequence of the
constraints imposed during training. Generative models are not optimized
solely to reproduce the empirical pixel distribution of real images.
Instead, their training objectives typically favor perceptual sharpness,
structural coherence, realistic texture, and consistency between local
details and global image content.

These constraints may introduce a statistical bias toward image structures
that are particularly salient to multiscale wavelet filters. In particular,
the penalization of blurred or weakly defined structures may reinforce
edges, contours, and local contrast variations, thereby increasing the
first-order scattering coefficients. At the same time, objectives that
encourage fine-scale details to remain consistent with larger semantic and
geometric structures may strengthen the coupling between spatial scales,
leading to larger second-order scattering coefficients.  These operations can produce images that are visually plausible but whose frequency statistics differ systematically from those of real images. Such spectral discrepancies have been repeatedly observed in generated imagery, including both excesses and structured inconsistencies in particular frequency band \cite{FourierDiscrepancies}.

Under this interpretation, the higher scattering response of generated
images is not simply a consequence of increased signal energy. Rather, it reflects a systematic amplification of perceptually relevant and multiscale-coherent structures induced by the training objective. The fact
that the patch-level coefficient of variation remains comparable, or is
slightly lower, for generated images further suggests that this effect is
approximately multiplicative: both the mean response and its absolute
spatial variability increase, while the relative variability remains
largely unchanged.

\subsubsection{Why is CLIP particularly sensitive to mesoscale statistics?}

A second question raised by our experiments concerns why CLIP appears to be particularly sensitive to mesoscale image statistics. One possible explanation lies in the contrastive objective used during pretraining. Positive image--text pairs are typically subjected to image transformations while preserving their semantic correspondence, so that the learned representation is encouraged to remain stable under changes that alter local image appearance without modifying the underlying content. Interestingly, the AI--human separation observed in our experiments is remarkably robust to several transformations of this kind, including cropping, resizing, grayscale conversion, and moderate degradation.

Under this perspective, the discriminative signal is unlikely to depend on isolated pixels or highly localized image structures. Instead, it should reside in statistical regularities that survive such transformations while remaining sufficiently informative to characterize the image. Mesoscale scattering statistics possess precisely these properties: they aggregate information across multiple spatial scales, are stable to local deformations, and retain structured information about interactions between image features at different scales.

We therefore hypothesize that the contrastive learning objective may favor visual cues with invariance properties similar to those captured by scattering representations. This does not imply that CLIP explicitly computes scattering-like features, but rather that distributed multiscale statistics may constitute a particularly effective form of visual evidence for a representation that must remain stable under substantial variation of the input. The robustness results reported in this work are consistent with this interpretation, although establishing a direct causal link between CLIP's training procedure and its sensitivity to mesoscale statistics would require dedicated experiments.

\subsection{Limitations}

The objective of the present work is not to provide an exhaustive characterization of every factor contributing to the separation between human and AI-generated paintings, but rather to investigate whether this phenomenon can be explained in terms of interpretable statistical image representations. Accordingly, our analysis follows a progressive strategy, moving from simple image statistics to increasingly expressive descriptors, with the aim of identifying the level of statistical complexity required to account for the observed geometry of the CLIP embedding space. Several important questions nevertheless remain open.

First, our analysis is restricted to CLIP representations. Although CLIP is one of the most influential vision foundation models, it remains to be established whether the statistical mechanisms identified in this work are specific to contrastive vision--language models or constitute a more general property of modern artificial vision systems. Extending the same methodology to self-supervised vision transformers and other foundation models represents a natural direction for future research.

Second, our experiments focus exclusively on artistic images. This choice is motivated by the central role of human perception in artistic evaluation and by the particularly clear separation observed in this domain. Whether similar statistical mechanisms govern the distinction between human and AI-generated natural photographs remains an open question.

Finally, although the robustness experiments rule out several simple explanations based on color, image quality, or preprocessing artifacts, they do not isolate every possible source of statistical regularity introduced by image-generation pipelines. For example, hidden generator-specific signatures or watermarking mechanisms cannot be completely excluded. We consider such explanations unlikely to account for the phenomenon in its entirety, given the distributed multiscale statistical differences revealed by the scattering analysis and the consistency of the observed behavior across different experimental settings. Nevertheless, explicitly disentangling these effects would further strengthen the interpretation proposed in this work.

These limitations do not alter the main empirical conclusions of the study: interpretable mesoscale statistics capture a meaningful component of the intrinsic geometry of the CLIP embedding space, but provide only a partial account of the observed separation. Rather than closing the problem, the present results identify the level at which part of the relevant visual information becomes accessible and define the scope for further investigation.

\section{Conclusion}
\label{sec:conclusion}
We investigated the spontaneous separation between human and AI-generated
paintings in the CLIP embedding space. Through a progressive analysis of
increasingly expressive interpretable image representations, we showed that
simple image statistics and local gradient descriptors account for only a
limited part of the observed geometry, whereas multiscale scattering
descriptors provide a substantially more informative statistical
representation. Scattering, nevertheless, provides only a partial account of
the phenomenon: it captures a meaningful component of the visual information
associated with the dominant principal directions, but a large part of 
their statistical structure remains unexplained.

A natural question is therefore what kind of interpretable representation could capture the statistical structure that remains unexplained. Scattering provides a multiscale description based on localized wavelet interactions and is conceptually close to the hierarchical processing characteristic of convolutional representations. CLIP, however, relies on a Vision Transformer, whose attention-based architecture can encode long-range and content-dependent interactions that are not naturally described within this framework. Developing interpretable statistical descriptors that more closely reflect attention-based processing may therefore provide access to aspects of the CLIP representation that are not captured by scattering, and constitutes an important direction for future investigation.

An equally important direction concerns the generality of the observed phenomenon. The present study focuses on CLIP because of its widespread adoption and well-established embedding space. Whether similar statistical mechanisms emerge in other vision foundation models, including self-supervised and purely visual architectures, remains an important open question. Comparative analyses across different representation learning paradigms may help distinguish properties that are specific to CLIP from more general characteristics of artificial vision.

More broadly, our results call into question an assumption that is often made implicitly throughout the computer vision community: namely, that modern vision foundation models extract visual information that is broadly aligned with human perception. The systematic discrepancies observed in this work suggest that this assumption deserves closer experimental scrutiny rather than being taken for granted.

This observation also has potential consequences for the interpretation of feature-based evaluation metrics. Measures such as FID and related perceptual distances rely on representations extracted by deep vision models and are therefore meaningful only insofar as these representations reflect perceptually relevant image differences. If artificial and human vision rely on systematically different statistical evidence, such metrics may reflect biases that do not necessarily coincide with human visual judgment.

Finally, we believe that these questions are particularly relevant in the artistic domain, where visual perception constitutes the primary medium of evaluation. If humans and artificial systems rely on different visual evidence, then agreement in semantic understanding does not necessarily imply agreement in aesthetic judgment. Rather than assuming that artificial vision and human perception converge toward a common notion of visual quality, future research should seek to understand where they coincide, where they diverge, and what these differences reveal about both natural and artificial intelligence.

\subsection*{Declarations}

\begin{itemize}
\item Conflict of interest. The author declares no conflict of interest. 
\item Data and code availability. Code, embeddings and precomputed features are available at the gitbub repository \href{https://github.com/asperti/Interpreting_Human_AI_separability}{Interpreting\_Human\_AI\_separability}.
\end{itemize}

%\bibliography{aesthetic,background}

\begin{thebibliography}{31}
\ifx \bisbn   \undefined \def \bisbn  #1{ISBN #1}\fi
\ifx \binits  \undefined \def \binits#1{#1}\fi
\ifx \bauthor  \undefined \def \bauthor#1{#1}\fi
\ifx \batitle  \undefined \def \batitle#1{#1}\fi
\ifx \bjtitle  \undefined \def \bjtitle#1{#1}\fi
\ifx \bvolume  \undefined \def \bvolume#1{\textbf{#1}}\fi
\ifx \byear  \undefined \def \byear#1{#1}\fi
\ifx \bissue  \undefined \def \bissue#1{#1}\fi
\ifx \bfpage  \undefined \def \bfpage#1{#1}\fi
\ifx \blpage  \undefined \def \blpage #1{#1}\fi
\ifx \burl  \undefined \def \burl#1{\textsf{#1}}\fi
\ifx \doiurl  \undefined \def \doiurl#1{\url{https://doi.org/#1}}\fi
\ifx \betal  \undefined \def \betal{\textit{et al.}}\fi
\ifx \binstitute  \undefined \def \binstitute#1{#1}\fi
\ifx \binstitutionaled  \undefined \def \binstitutionaled#1{#1}\fi
\ifx \bctitle  \undefined \def \bctitle#1{#1}\fi
\ifx \beditor  \undefined \def \beditor#1{#1}\fi
\ifx \bpublisher  \undefined \def \bpublisher#1{#1}\fi
\ifx \bbtitle  \undefined \def \bbtitle#1{#1}\fi
\ifx \bedition  \undefined \def \bedition#1{#1}\fi
\ifx \bseriesno  \undefined \def \bseriesno#1{#1}\fi
\ifx \blocation  \undefined \def \blocation#1{#1}\fi
\ifx \bsertitle  \undefined \def \bsertitle#1{#1}\fi
\ifx \bsnm \undefined \def \bsnm#1{#1}\fi
\ifx \bsuffix \undefined \def \bsuffix#1{#1}\fi
\ifx \bparticle \undefined \def \bparticle#1{#1}\fi
\ifx \barticle \undefined \def \barticle#1{#1}\fi
\bibcommenthead
\ifx \bconfdate \undefined \def \bconfdate #1{#1}\fi
\ifx \botherref \undefined \def \botherref #1{#1}\fi
\ifx \url \undefined \def \url#1{\textsf{#1}}\fi
\ifx \bchapter \undefined \def \bchapter#1{#1}\fi
\ifx \bbook \undefined \def \bbook#1{#1}\fi
\ifx \bcomment \undefined \def \bcomment#1{#1}\fi
\ifx \oauthor \undefined \def \oauthor#1{#1}\fi
\ifx \citeauthoryear \undefined \def \citeauthoryear#1{#1}\fi
\ifx \endbibitem  \undefined \def \endbibitem {}\fi
\ifx \bconflocation  \undefined \def \bconflocation#1{#1}\fi
\ifx \arxivurl  \undefined \def \arxivurl#1{\textsf{#1}}\fi
\csname PreBibitemsHook\endcsname

%%% 1
\bibitem[\protect\citeauthoryear{Liao et~al.}{2022}]{artbench_dataset}
\begin{botherref}
\oauthor{\bsnm{Liao}, \binits{P.}},
\oauthor{\bsnm{Li}, \binits{X.}},
\oauthor{\bsnm{Liu}, \binits{X.}},
\oauthor{\bsnm{Keutzer}, \binits{K.}}:
The artbench dataset: Benchmarking generative models with artworks.
CoRR
\textbf{abs/2206.11404}
(2022)
\doiurl{10.48550/ARXIV.2206.11404}
{\href{https://arxiv.org/abs/2206.11404}{{2206.11404}}}
\end{botherref}
\endbibitem

%%% 2
\bibitem[\protect\citeauthoryear{Asperti et~al.}{2026}]{Art_CLIPs_eyes}
\begin{barticle}
\bauthor{\bsnm{Asperti}, \binits{A.}},
\bauthor{\bsnm{Dess\`{\i}}, \binits{L.}},
\bauthor{\bsnm{Wu}, \binits{N.}}:
\batitle{Art through clip’s eyes}.
\bjtitle{J. Comput. Cult. Herit.}
(\byear{2026})
\doiurl{10.1145/3812546} .
\bcomment{Just Accepted}
\end{barticle}
\endbibitem

%%% 3
\bibitem[\protect\citeauthoryear{Asperti et~al.}{2025}]{AI-pastiche}
\begin{barticle}
\bauthor{\bsnm{Asperti}, \binits{A.}},
\bauthor{\bsnm{George}, \binits{F.}},
\bauthor{\bsnm{Marras}, \binits{T.}},
\bauthor{\bsnm{Stricescu}, \binits{R.C.}},
\bauthor{\bsnm{Zanotti}, \binits{F.}}:
\batitle{A critical assessment of modern generative models’ ability to replicate artistic styles}.
\bjtitle{Big Data and Cognitive Computing}
\bvolume{9}(\bissue{9}),
\bfpage{1}--\blpage{26}
(\byear{2025})
\doiurl{10.3390/bdcc9090231}
\end{barticle}
\endbibitem

%%% 4
\bibitem[\protect\citeauthoryear{of~Art}{2024}]{nga_real_dataset}
\begin{botherref}
\oauthor{\bsnm{Art}, \binits{N.G.}}:
National Gallery of Art Open Data Program.
Accessed: 2024-01-29
(2024).
\url{https://www.nga.gov/open-access-images/open-data.html}
\end{botherref}
\endbibitem

%%% 5
\bibitem[\protect\citeauthoryear{Radford et~al.}{2021}]{CLIP}
\begin{bchapter}
\bauthor{\bsnm{Radford}, \binits{A.}},
\bauthor{\bsnm{Kim}, \binits{J.W.}},
\bauthor{\bsnm{Hallacy}, \binits{C.}},
\bauthor{\bsnm{Ramesh}, \binits{A.}},
\bauthor{\bsnm{Goh}, \binits{G.}},
\bauthor{\bsnm{Agarwal}, \binits{S.}},
\bauthor{\bsnm{Sastry}, \binits{G.}},
\bauthor{\bsnm{Askell}, \binits{A.}},
\bauthor{\bsnm{Mishkin}, \binits{P.}},
\bauthor{\bsnm{Clark}, \binits{J.}},
\bauthor{\bsnm{Krueger}, \binits{G.}},
\bauthor{\bsnm{Sutskever}, \binits{I.}}:
\bctitle{Learning transferable visual models from natural language supervision}.
In: \beditor{\bsnm{Meila}, \binits{M.}},
\beditor{\bsnm{Zhang}, \binits{T.}} (eds.)
\bbtitle{Proceedings of the 38th International Conference on Machine Learning, {ICML} 2021, 18-24 July 2021, Virtual Event}.
\bsertitle{Proceedings of Machine Learning Research},
vol. \bseriesno{139},
pp. \bfpage{8748}--\blpage{8763}.
\bpublisher{{PMLR}},
\blocation{\hspace{0pt}}
(\byear{2021}).
\bcomment{Accessed: 2025-02-13}.
\burl{http://proceedings.mlr.press/v139/radford21a.html}
\end{bchapter}
\endbibitem

%%% 6
\bibitem[\protect\citeauthoryear{Bhalla et~al.}{2024}]{splice24}
\begin{bchapter}
\bauthor{\bsnm{Bhalla}, \binits{U.}},
\bauthor{\bsnm{Oesterling}, \binits{A.}},
\bauthor{\bsnm{Srinivas}, \binits{S.}},
\bauthor{\bsnm{Calmon}, \binits{F.P.}},
\bauthor{\bsnm{Lakkaraju}, \binits{H.}}:
\bctitle{Interpreting {CLIP} with sparse linear concept embeddings (splice)}.
In: \beditor{\bsnm{Globersons}, \binits{A.}},
\beditor{\bsnm{Mackey}, \binits{L.}},
\beditor{\bsnm{Belgrave}, \binits{D.}},
\beditor{\bsnm{Fan}, \binits{A.}},
\beditor{\bsnm{Paquet}, \binits{U.}},
\beditor{\bsnm{Tomczak}, \binits{J.M.}},
\beditor{\bsnm{Zhang}, \binits{C.}} (eds.)
\bbtitle{Advances in Neural Information Processing Systems 37: Annual Conference on Neural Information Processing Systems 2024, NeurIPS 2024, Vancouver, BC, Canada, December 10 - 15, 2024}
(\byear{2024}).
\burl{http://papers.nips.cc/paper\_files/paper/2024/hash/996bef37d8a638f37bdfcac2789e835d-Abstract-Conference.html}
\end{bchapter}
\endbibitem

%%% 7
\bibitem[\protect\citeauthoryear{Goh et~al.}{2021}]{goh2021multimodal}
\begin{barticle}
\bauthor{\bsnm{Goh}, \binits{G.}},
\bauthor{\bsnm{†}, \binits{N.C.}},
\bauthor{\bsnm{†}, \binits{C.V.}},
\bauthor{\bsnm{Carter}, \binits{S.}},
\bauthor{\bsnm{Petrov}, \binits{M.}},
\bauthor{\bsnm{Schubert}, \binits{L.}},
\bauthor{\bsnm{Radford}, \binits{A.}},
\bauthor{\bsnm{Olah}, \binits{C.}}:
\batitle{Multimodal neurons in artificial neural networks}.
\bjtitle{Distill}
(\byear{2021})
\doiurl{10.23915/distill.00030} .
\bcomment{https://distill.pub/2021/multimodal-neurons}
\end{barticle}
\endbibitem

%%% 8
\bibitem[\protect\citeauthoryear{Gandelsman et~al.}{2024}]{CLIP_via_text24}
\begin{bchapter}
\bauthor{\bsnm{Gandelsman}, \binits{Y.}},
\bauthor{\bsnm{Efros}, \binits{A.A.}},
\bauthor{\bsnm{Steinhardt}, \binits{J.}}:
\bctitle{Interpreting clip's image representation via text-based decomposition}.
In: \bbtitle{The Twelfth International Conference on Learning Representations, {ICLR} 2024, Vienna, Austria, May 7-11, 2024}.
\bpublisher{OpenReview.net},
\blocation{\hspace{0pt}}
(\byear{2024}).
\burl{https://openreview.net/forum?id=5Ca9sSzuDp}
\end{bchapter}
\endbibitem

%%% 9
\bibitem[\protect\citeauthoryear{Oikarinen and Weng}{2023}]{CLIP-dissect}
\begin{bchapter}
\bauthor{\bsnm{Oikarinen}, \binits{T.P.}},
\bauthor{\bsnm{Weng}, \binits{T.}}:
\bctitle{Clip-dissect: Automatic description of neuron representations in deep vision networks}.
In: \bbtitle{The Eleventh International Conference on Learning Representations, {ICLR} 2023, Kigali, Rwanda, May 1-5, 2023}.
\bpublisher{OpenReview.net},
\blocation{\hspace{0pt}}
(\byear{2023}).
\burl{https://openreview.net/forum?id=iPWiwWHc1V}
\end{bchapter}
\endbibitem

%%% 10
\bibitem[\protect\citeauthoryear{Wang et~al.}{2020}]{easy_to_spot20}
\begin{bchapter}
\bauthor{\bsnm{Wang}, \binits{S.}},
\bauthor{\bsnm{Wang}, \binits{O.}},
\bauthor{\bsnm{Zhang}, \binits{R.}},
\bauthor{\bsnm{Owens}, \binits{A.}},
\bauthor{\bsnm{Efros}, \binits{A.A.}}:
\bctitle{Cnn-generated images are surprisingly easy to spot... for now}.
In: \bbtitle{2020 {IEEE/CVF} Conference on Computer Vision and Pattern Recognition, {CVPR} 2020, Seattle, WA, USA, June 13-19, 2020},
pp. \bfpage{8692}--\blpage{8701}.
\bpublisher{Computer Vision Foundation / {IEEE}},
\blocation{\hspace{0pt}}
(\byear{2020}).
\doiurl{10.1109/CVPR42600.2020.00872} .
\burl{https://openaccess.thecvf.com/content\_CVPR\_2020/html/Wang\_CNN-Generated\_Images\_Are\_Surprisingly\_Easy\_to\_Spot...\_for\_Now\_CVPR\_2020\_paper.html}
\end{bchapter}
\endbibitem

%%% 11
\bibitem[\protect\citeauthoryear{Gragnaniello et~al.}{2021}]{GragnanielloCMP21}
\begin{bchapter}
\bauthor{\bsnm{Gragnaniello}, \binits{D.}},
\bauthor{\bsnm{Cozzolino}, \binits{D.}},
\bauthor{\bsnm{Marra}, \binits{F.}},
\bauthor{\bsnm{Poggi}, \binits{G.}},
\bauthor{\bsnm{Verdoliva}, \binits{L.}}:
\bctitle{Are {GAN} generated images easy to detect? {A} critical analysis of the state-of-the-art}.
In: \bbtitle{2021 {IEEE} International Conference on Multimedia and Expo, {ICME} 2021, Shenzhen, China, July 5-9, 2021},
pp. \bfpage{1}--\blpage{6}.
\bpublisher{{IEEE}},
\blocation{\hspace{0pt}}
(\byear{2021}).
\doiurl{10.1109/ICME51207.2021.9428429} .
\burl{https://doi.org/10.1109/ICME51207.2021.9428429}
\end{bchapter}
\endbibitem

%%% 12
\bibitem[\protect\citeauthoryear{Cozzolino et~al.}{2024}]{Raising_the_bar24}
\begin{bchapter}
\bauthor{\bsnm{Cozzolino}, \binits{D.}},
\bauthor{\bsnm{Poggi}, \binits{G.}},
\bauthor{\bsnm{Corvi}, \binits{R.}},
\bauthor{\bsnm{Nie{\ss}ner}, \binits{M.}},
\bauthor{\bsnm{Verdoliva}, \binits{L.}}:
\bctitle{Raising the bar of ai-generated image detection with {CLIP}}.
In: \bbtitle{{IEEE/CVF} Conference on Computer Vision and Pattern Recognition, {CVPR} 2024 - Workshops, Seattle, WA, USA, June 17-18, 2024},
pp. \bfpage{4356}--\blpage{4366}.
\bpublisher{{IEEE}},
\blocation{\hspace{0pt}}
(\byear{2024}).
\doiurl{10.1109/CVPRW63382.2024.00439} .
\burl{https://doi.org/10.1109/CVPRW63382.2024.00439}
\end{bchapter}
\endbibitem

%%% 13
\bibitem[\protect\citeauthoryear{Moskowitz et~al.}{2024}]{DetectingAI}
\begin{botherref}
\oauthor{\bsnm{Moskowitz}, \binits{A.G.}},
\oauthor{\bsnm{Gaona}, \binits{T.}},
\oauthor{\bsnm{Peterson}, \binits{J.}}:
Detecting ai-generated images via {CLIP}.
CoRR
\textbf{abs/2404.08788}
(2024)
\doiurl{10.48550/ARXIV.2404.08788}
{\href{https://arxiv.org/abs/2404.08788}{{2404.08788}}}
\end{botherref}
\endbibitem

%%% 14
\bibitem[\protect\citeauthoryear{Gaintseva et~al.}{2024}]{Improving_interpretability24}
\begin{botherref}
\oauthor{\bsnm{Gaintseva}, \binits{T.}},
\oauthor{\bsnm{Kushnareva}, \binits{L.}},
\oauthor{\bsnm{Magai}, \binits{G.}},
\oauthor{\bsnm{Piontkovskaya}, \binits{I.}},
\oauthor{\bsnm{Nikolenko}, \binits{S.I.}},
\oauthor{\bsnm{Benning}, \binits{M.}},
\oauthor{\bsnm{Barannikov}, \binits{S.}},
\oauthor{\bsnm{Slabaugh}, \binits{G.G.}}:
Improving interpretability and robustness for the detection of ai-generated images.
CoRR
\textbf{abs/2406.15035}
(2024)
\doiurl{10.48550/ARXIV.2406.15035}
{\href{https://arxiv.org/abs/2406.15035}{{2406.15035}}}
\end{botherref}
\endbibitem

%%% 15
\bibitem[\protect\citeauthoryear{Tuytelaars and Mikolajczyk}{2007}]{tuytelaars08local}
\begin{barticle}
\bauthor{\bsnm{Tuytelaars}, \binits{T.}},
\bauthor{\bsnm{Mikolajczyk}, \binits{K.}}:
\batitle{Local invariant feature detectors: {A} survey}.
\bjtitle{Found. Trends Comput. Graph. Vis.}
\bvolume{3}(\bissue{3}),
\bfpage{177}--\blpage{280}
(\byear{2007})
\doiurl{10.1561/0600000017}
\end{barticle}
\endbibitem

%%% 16
\bibitem[\protect\citeauthoryear{Vigneron and Sylvie}{2010}]{statistics_for_images}
\begin{barticle}
\bauthor{\bsnm{Vigneron}, \binits{V.}},
\bauthor{\bsnm{Sylvie}, \binits{L.}}:
\batitle{Statistics for image processing}.
\bjtitle{Pattern Recognition Letters}
\bvolume{31},
\bfpage{2191}
(\byear{2010})
\doiurl{10.1016/j.patrec.2010.07.003}
\end{barticle}
\endbibitem

%%% 17
\bibitem[\protect\citeauthoryear{Lowe}{2004}]{lowe2004distinctive}
\begin{barticle}
\bauthor{\bsnm{Lowe}, \binits{D.G.}}:
\batitle{Distinctive image features from scale-invariant keypoints}.
\bjtitle{Int. J. Comput. Vis.}
\bvolume{60}(\bissue{2}),
\bfpage{91}--\blpage{110}
(\byear{2004})
\doiurl{10.1023/B:VISI.0000029664.99615.94}
\end{barticle}
\endbibitem

%%% 18
\bibitem[\protect\citeauthoryear{Dalal and Triggs}{2005}]{dalal2005histograms}
\begin{bchapter}
\bauthor{\bsnm{Dalal}, \binits{N.}},
\bauthor{\bsnm{Triggs}, \binits{B.}}:
\bctitle{Histograms of oriented gradients for human detection}.
In: \bbtitle{2005 {IEEE} Computer Society Conference on Computer Vision and Pattern Recognition {(CVPR} 2005), 20-26 June 2005, San Diego, CA, {USA}},
pp. \bfpage{886}--\blpage{893}.
\bpublisher{{IEEE} Computer Society},
\blocation{\hspace{0pt}}
(\byear{2005}).
\doiurl{10.1109/CVPR.2005.177} .
\burl{https://doi.org/10.1109/CVPR.2005.177}
\end{bchapter}
\endbibitem

%%% 19
\bibitem[\protect\citeauthoryear{Portilla and Simoncelli}{2000}]{portilla2000parametric}
\begin{barticle}
\bauthor{\bsnm{Portilla}, \binits{J.}},
\bauthor{\bsnm{Simoncelli}, \binits{E.P.}}:
\batitle{A parametric texture model based on joint statistics of complex wavelet coefficients}.
\bjtitle{Int. J. Comput. Vis.}
\bvolume{40}(\bissue{1}),
\bfpage{49}--\blpage{70}
(\byear{2000})
\doiurl{10.1023/A:1026553619983}
\end{barticle}
\endbibitem

%%% 20
\bibitem[\protect\citeauthoryear{Simoncelli and Olshausen}{2001}]{simoncelli2001natural}
\begin{barticle}
\bauthor{\bsnm{Simoncelli}, \binits{E.P.}},
\bauthor{\bsnm{Olshausen}, \binits{B.A.}}:
\batitle{Natural image statistics and neural representation}.
\bjtitle{Annual review of neuroscience}
\bvolume{24}(\bissue{1}),
\bfpage{1193}--\blpage{1216}
(\byear{2001})
\end{barticle}
\endbibitem

%%% 21
\bibitem[\protect\citeauthoryear{Mallat}{2011}]{mallat2011group}
\begin{botherref}
\oauthor{\bsnm{Mallat}, \binits{S.}}:
Group invariant scattering.
CoRR
\textbf{abs/1101.2286}
(2011)
{\href{https://arxiv.org/abs/1101.2286}{{1101.2286}}}
\end{botherref}
\endbibitem

%%% 22
\bibitem[\protect\citeauthoryear{Bruna and Mallat}{2013}]{bruna2013invariant}
\begin{barticle}
\bauthor{\bsnm{Bruna}, \binits{J.}},
\bauthor{\bsnm{Mallat}, \binits{S.}}:
\batitle{Invariant scattering convolution networks}.
\bjtitle{{IEEE} Trans. Pattern Anal. Mach. Intell.}
\bvolume{35}(\bissue{8}),
\bfpage{1872}--\blpage{1886}
(\byear{2013})
\doiurl{10.1109/TPAMI.2012.230}
\end{barticle}
\endbibitem

%%% 23
\bibitem[\protect\citeauthoryear{Sifre and Mallat}{2013}]{sifre2013rotation}
\begin{bchapter}
\bauthor{\bsnm{Sifre}, \binits{L.}},
\bauthor{\bsnm{Mallat}, \binits{S.}}:
\bctitle{Rotation, scaling and deformation invariant scattering for texture discrimination}.
In: \bbtitle{2013 {IEEE} Conference on Computer Vision and Pattern Recognition, Portland, OR, USA, June 23-28, 2013},
pp. \bfpage{1233}--\blpage{1240}.
\bpublisher{{IEEE} Computer Society},
\blocation{\hspace{0pt}}
(\byear{2013}).
\doiurl{10.1109/CVPR.2013.163} .
\burl{https://doi.org/10.1109/CVPR.2013.163}
\end{bchapter}
\endbibitem

%%% 24
\bibitem[\protect\citeauthoryear{Erhan et~al.}{2009}]{erhan2009visualizing}
\begin{botherref}
\oauthor{\bsnm{Erhan}, \binits{D.}},
\oauthor{\bsnm{Bengio}, \binits{Y.}},
\oauthor{\bsnm{Courville}, \binits{A.}},
\oauthor{\bsnm{Vincent}, \binits{P.}}:
Visualizing higher-layer features of a deep network.
Technical Report 1341,
University of Montreal, Canada
(June 2009)
\end{botherref}
\endbibitem

%%% 25
\bibitem[\protect\citeauthoryear{Yosinski et~al.}{2015}]{yosinski2015understanding}
\begin{botherref}
\oauthor{\bsnm{Yosinski}, \binits{J.}},
\oauthor{\bsnm{Clune}, \binits{J.}},
\oauthor{\bsnm{Nguyen}, \binits{A.M.}},
\oauthor{\bsnm{Fuchs}, \binits{T.J.}},
\oauthor{\bsnm{Lipson}, \binits{H.}}:
Understanding neural networks through deep visualization.
CoRR
\textbf{abs/1506.06579}
(2015)
{\href{https://arxiv.org/abs/1506.06579}{{1506.06579}}}
\end{botherref}
\endbibitem

%%% 26
\bibitem[\protect\citeauthoryear{Mahendran and Vedaldi}{2015}]{Vevaldi15}
\begin{bchapter}
\bauthor{\bsnm{Mahendran}, \binits{A.}},
\bauthor{\bsnm{Vedaldi}, \binits{A.}}:
\bctitle{Understanding deep image representations by inverting them}.
In: \bbtitle{{IEEE} Conference on Computer Vision and Pattern Recognition, {CVPR} 2015, Boston, MA, USA, June 7-12, 2015},
pp. \bfpage{5188}--\blpage{5196}.
\bpublisher{{IEEE} Computer Society},
\blocation{\hspace{0pt}}
(\byear{2015}).
\doiurl{10.1109/CVPR.2015.7299155} .
\burl{https://doi.org/10.1109/CVPR.2015.7299155}
\end{bchapter}
\endbibitem

%%% 27
\bibitem[\protect\citeauthoryear{Dosovitskiy and Brox}{2016}]{inverting_CNN}
\begin{bchapter}
\bauthor{\bsnm{Dosovitskiy}, \binits{A.}},
\bauthor{\bsnm{Brox}, \binits{T.}}:
\bctitle{Inverting visual representations with convolutional networks}.
In: \bbtitle{2016 {IEEE} Conference on Computer Vision and Pattern Recognition, {CVPR} 2016, Las Vegas, NV, USA, June 27-30, 2016},
pp. \bfpage{4829}--\blpage{4837}.
\bpublisher{{IEEE} Computer Society},
\blocation{\hspace{0pt}}
(\byear{2016}).
\burl{https://doi.org/10.1109/CVPR.2016.522}
\end{bchapter}
\endbibitem

%%% 28
\bibitem[\protect\citeauthoryear{Szegedy et~al.}{2014}]{Intriguing2013}
\begin{bchapter}
\bauthor{\bsnm{Szegedy}, \binits{C.}},
\bauthor{\bsnm{Zaremba}, \binits{W.}},
\bauthor{\bsnm{Sutskever}, \binits{I.}},
\bauthor{\bsnm{Bruna}, \binits{J.}},
\bauthor{\bsnm{Erhan}, \binits{D.}},
\bauthor{\bsnm{Goodfellow}, \binits{I.J.}},
\bauthor{\bsnm{Fergus}, \binits{R.}}:
\bctitle{Intriguing properties of neural networks}.
In: \beditor{\bsnm{Bengio}, \binits{Y.}},
\beditor{\bsnm{LeCun}, \binits{Y.}} (eds.)
\bbtitle{2nd International Conference on Learning Representations, {ICLR} 2014, Banff, AB, Canada, April 14-16, 2014, Conference Track Proceedings}
(\byear{2014}).
\burl{http://arxiv.org/abs/1312.6199}
\end{bchapter}
\endbibitem

%%% 29
\bibitem[\protect\citeauthoryear{Goodfellow et~al.}{2015}]{goodfellow2014explaining}
\begin{bchapter}
\bauthor{\bsnm{Goodfellow}, \binits{I.J.}},
\bauthor{\bsnm{Shlens}, \binits{J.}},
\bauthor{\bsnm{Szegedy}, \binits{C.}}:
\bctitle{Explaining and harnessing adversarial examples}.
In: \beditor{\bsnm{Bengio}, \binits{Y.}},
\beditor{\bsnm{LeCun}, \binits{Y.}} (eds.)
\bbtitle{3rd International Conference on Learning Representations, {ICLR} 2015, San Diego, CA, USA, May 7-9, 2015, Conference Track Proceedings}
(\byear{2015}).
\burl{http://arxiv.org/abs/1412.6572}
\end{bchapter}
\endbibitem

%%% 30
\bibitem[\protect\citeauthoryear{Fu et~al.}{2025}]{WikiArt25}
\begin{barticle}
\bauthor{\bsnm{Fu}, \binits{T.}},
\bauthor{\bsnm{Conde}, \binits{J.}},
\bauthor{\bsnm{Martínez}, \binits{G.}},
\bauthor{\bsnm{Reviriego}, \binits{P.}},
\bauthor{\bsnm{Merino-Gómez}, \binits{E.}},
\bauthor{\bsnm{Moral}, \binits{F.}}:
\batitle{Artificial intelligence and misinformation in art: Can vision language models judge the hand or the machine behind the canvas?}
\bjtitle{CoRR}
\bvolume{abs/2508.01408},
\bfpage{1}--\blpage{18}
(\byear{2025})
{\href{https://arxiv.org/abs/2508.01408}{{2508.01408}}}
\end{barticle}
\endbibitem

%%% 31
\bibitem[\protect\citeauthoryear{Dzanic et~al.}{2020}]{FourierDiscrepancies}
\begin{bchapter}
\bauthor{\bsnm{Dzanic}, \binits{T.}},
\bauthor{\bsnm{Shah}, \binits{K.}},
\bauthor{\bsnm{Witherden}, \binits{F.D.}}:
\bctitle{Fourier spectrum discrepancies in deep network generated images}.
In: \beditor{\bsnm{Larochelle}, \binits{H.}},
\beditor{\bsnm{Ranzato}, \binits{M.}},
\beditor{\bsnm{Hadsell}, \binits{R.}},
\beditor{\bsnm{Balcan}, \binits{M.}},
\beditor{\bsnm{Lin}, \binits{H.}} (eds.)
\bbtitle{Advances in Neural Information Processing Systems 33: Annual Conference on Neural Information Processing Systems 2020, NeurIPS 2020, December 6-12, 2020, Virtual}
(\byear{2020}).
\burl{https://proceedings.neurips.cc/paper/2020/hash/1f8d87e1161af68b81bace188a1ec624-Abstract.html}
\end{bchapter}
\endbibitem

\end{thebibliography}
%\input{output.bbl}

\end{document}